\documentclass[10pt,journal,compsoc]{IEEEtran}

\ifCLASSOPTIONcompsoc
  \usepackage[nocompress]{cite}
\else
  \usepackage{cite}
\fi
\usepackage{graphicx}
\usepackage{amsmath,amssymb}
\usepackage{color}
\usepackage{multirow}
\usepackage{booktabs}
\usepackage[T1]{fontenc}
\usepackage{subfigure}
\usepackage{float}
\usepackage{color}
\usepackage{setspace}
\usepackage{url}
\usepackage{makecell}
\usepackage{cleveref}

\begin{document}
%
\title{Uncertainty-boosted \\ Robust Video Activity Anticipation}

\author{Zhaobo Qi, Shuhui Wang,~\IEEEmembership{Member,~IEEE}, Weigang Zhang,~\IEEEmembership{Member,~IEEE} \\ and Qingming Huang,~\IEEEmembership{Fellow,~IEEE}
\IEEEcompsocitemizethanks{
	\IEEEcompsocthanksitem Corresponding author: Shuhui Wang and Weigang Zhang.
	\IEEEcompsocthanksitem  Z. Qi and W. Zhang is with the School of Computer Science and Technology, Harbin Institute of Technology, Weihai 264209, China. \protect\\
	E-mail: qizb@hit.edu.cn, wgzhang@hit.edu.cn.
	\IEEEcompsocthanksitem  S. Wang is with the Key Laboratory of Intelligent Information Processing, Institute of Computing Technology, Chinese Academy of Sciences, Beijing 100190, China, and with Peng Cheng Laboratory, Shenzhen 518066, China. \protect\\
	E-mail: wangshuhui@ict.ac.cn.
	\IEEEcompsocthanksitem  Q. Huang is with the School of Computer Science and Technology, University of Chinese Academy of Sciences, Beijing 101408, China, with the Key Laboratory of Intelligent Information Processing, Institute of Computing Technology, Chinese Academy of Sciences, Beijing 100190, China, and with Peng Cheng Laboratory, Shenzhen 518066, China. \protect\\
	E-mail: qmhuang@ucas.ac.cn.}
} 


\markboth{Transactions on Pattern Analysis and Machine Intelligence}%
{Shell \MakeLowercase{\textit{et al.}}: Bare Demo of IEEEtran.cls for Computer Society Journals}

\IEEEtitleabstractindextext{%
	\begin{abstract}
		Video activity anticipation aims to predict what will happen in the future, embracing a broad application prospect ranging from robot vision and autonomous driving. Despite the recent progress, the data uncertainty issue, reflected as the content evolution process and dynamic correlation in event labels, has been somehow ignored. 
		This reduces the model generalization ability and deep understanding on video content, leading to serious error accumulation and degraded performance. 
		In this paper, we address the uncertainty learning problem and propose an uncertainty-boosted robust video activity anticipation framework, which generates uncertainty values to indicate the credibility of the anticipation results. The uncertainty value is used to derive a temperature parameter in the softmax function to modulate the predicted target activity distribution. To guarantee the distribution adjustment, we construct a reasonable target activity label representation by incorporating the activity evolution from the temporal class correlation and the semantic relationship.
		Moreover, we quantify the uncertainty into relative values by comparing the uncertainty among sample pairs and their temporal-lengths. This relative strategy provides a more accessible way in uncertainty modeling than quantifying the absolute uncertainty values on the whole dataset.  	
		Experiments on multiple backbones and benchmarks show our framework achieves promising performance and better robustness/interpretability. Source codes are available at https://github.com/qzhb/UbRV2A.
	\end{abstract}
	\begin{IEEEkeywords}
		Video Activity Anticipation, Data Uncertainty, Relative Uncertainty Learning, Robustness.
\end{IEEEkeywords}}

\maketitle

\IEEEdisplaynontitleabstractindextext

%
\IEEEpeerreviewmaketitle

\IEEEraisesectionheading{\section{Introduction}\label{sec:introduction}}

The intelligent video analytics has been rapidly developed, and benefits applications such as human-robot interaction and autonomous driving~\cite{de2016online}. 
Among various types of video analytic tasks, video activity anticipation~\cite{koppula2015anticipating} receives increasing attention in the research community, which aims to predict what will happen in the future by anticipating the activity categories. Following~\cite{hutchinson2021video}, recent advances on this challenging task are roughly divided into generative families that perform target activity classification on the anticipated features~\cite{abu2018will,rulstm,srl,zatsarynna2021multi,camporese2021knowledge,fernando2021anticipating}, and non-generative families that directly generate anticipation results based on the observed features~\cite{qi2017predicting,furnari2017next,abu2019uncertainty,furnari2018leveraging,ke2019time}. 
Despite the promising performance, most work directly produces 
the anticipation output without any evidence support, leading to serious consequences such as unpredictable model behavior and low interpretability. 
Towards better robustness and trustability~\cite{blundell2015weight,gal2016dropout,kendall2017uncertainties}, the anticipation model is expected to produce the correct output with low uncertainty and erroneous outputs with high uncertainty, providing us with decision support and avoiding potential risks.

The key to achieving this goal is to model and manage the uncertainty within the video data and anticipation model. 
There are multiple potential sources of uncertainty involved in the three critical stages from the raw video data inputs recording the scenarios to the anticipation outputs generated by deep neural networks (DNNs), namely, the data acquisition process, the design and training of a DNN, and the inference of the DNN. 
Consequently, uncertainty in video anticipation output stems aleatorily and epistemically, {\it a.k.a} data and model uncertainty~\cite{hora1996aleatory, der2009aleatory}. 
The former indicates the inherent property of video data introduced during the data collection and annotation process, which cannot be eliminated even with increased data size. 
The latter, encapsulated in model parameters, encompasses the insufficient capability of model structure, training data coverage, errors in the training process, or other similar phenomena.

An intuitive countermeasure to alleviate uncertainty is to jointly model aleatoric and epistemic ones~\cite{kendall2017uncertainties}. However, as discussed in~\cite{kendall2017uncertainties, ayhan2022test}, in the big data application context, the performance depends largely on the influence of data uncertainty, while the contribution of model uncertainty appears to be negligible, {\it i.e.}, one can reduce model uncertainty with small effort using well-established techniques like Monte Carlo Dropout ~\cite{gal2016dropout, kendall2017uncertainties, depeweg2018decomposition}. The reason can be explained from two aspects. First, the assumptions on the hypothesis space (or model family) of deep learning models are large enough to capture fundamental facts from data~\cite{hullermeier2021aleatoric, gruber2023sources}. In particular, deep visual foundation models with billions of parameters exhibit high representation and anticipation ability, expressing the hypothesis space powerfully. Consequently, the model uncertainty becomes less significant when we impose weak assumptions on model structure and parameters, and it can be mitigated by simply enlarging the training data size. 	
Second, the data uncertainty that contributes into the learning process of DNNs can compensate for model uncertainty when the two are modeled independently~\cite{kendall2017uncertainties, ayhan2022test}. For instance, the data uncertainty and model uncertainty lead to similar a ranking of pixel-wise prediction confidence in semantic segmentation and depth estimation tasks~\cite{kendall2017uncertainties}. 
This indicates that there exists a certain level of equivalence between data uncertainty and model uncertainty, where similar relation can be found from classical machine learning theories such as the Reproductive Kernel Hilbert Space~(RKHS)~\cite{paulsen2016introduction} that connects the data distribution and model parameters for constructing the kernelized models.
Therefore, it is reasonable to prioritize the data uncertainty under real-world scenarios with large-scale data, to gain insight into special datasets and tasks towards improved model robustness.

\begin{figure}[t]
	\centering
	\subfigure[In each example, three video activity sequences with the same antecedent activity and different upcoming activities are given.{\label{motivation_example}}] 
	{\includegraphics[width=1\linewidth]{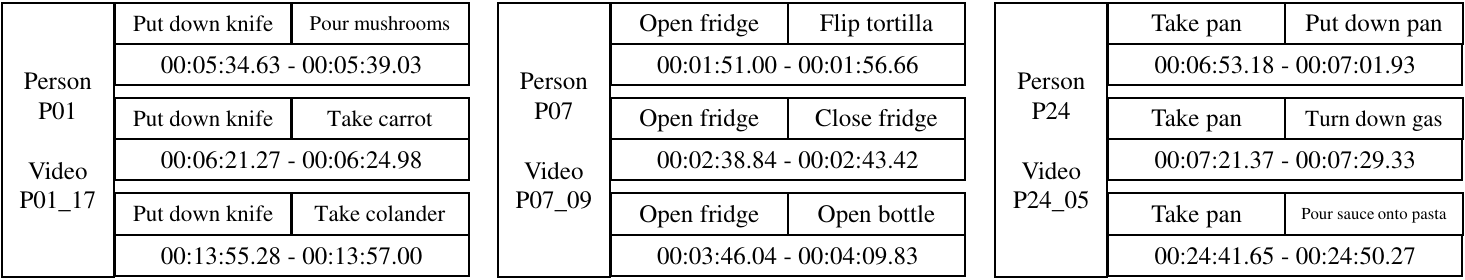}}
	\quad
	\subfigure[The number of video instances that contain the corresponding antecedent and subsequent activity classes. {\label{motivation_heatmap}}]
	{\includegraphics[width=0.95\linewidth]{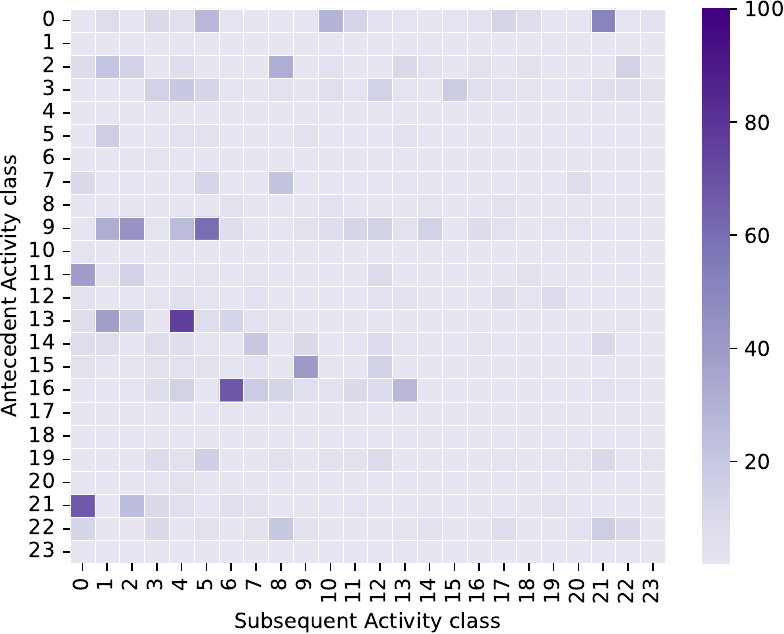}}
	\quad
	\caption{The activity video evolution uncertainty phenomenon on EPIC-KITCHENS-55 dataset. For clarity, partial classes are shown.}
	\label{fig_motivation}
\end{figure}

In the video activity anticipation task, diverse factors are involved to yield data uncertainty. For example, videos in the widely used EPIC-KITCHENS-55 dataset~\cite{damen2018scaling} are recorded by 32 individuals in 32 kitchens using adjustable head-mounted GoPro. In this process, the perspective of data recording and the status of actors bring about large data uncertainty in content expression. Then, each volunteer is asked to narrate the activities carried out, and the Amazon Mechanical Turk is used to employ ordinary users to further annotate these data. 
The preference of annotators and diversity in the understanding of the video sequences further lead to uncertainty in the activity labels. 
The uncertainty along the evolution process in the activity video streams is also worth exploring. 
As shown in Figure~\ref{motivation_example}, given videos with the same antecedent activity, even the same person in the same kitchen may take different subsequent activities\footnote{All sequences are sampled from the same video, and each video is recorded on the same person in the same kitchen at once.}. 
More clearly, for each pair of activity classes, we count the number of video instances that contain these two consecutive classes on EPIC-KITCHENS-55, and obtain a square matrix in Figure~\ref{motivation_heatmap}. 
Each row represents the distribution of one antecedent activity evolving into others, indicating that a plethora of activity categories may occur next from the same antecedent activity. Hence, the activity semantics evolution appears to be highly uncertain.

The data uncertainty seriously impedes the reliability of the anticipation model. 
Concretely, it results in poor generalization on samples following a flat distribution or activity categories with a large number of possible subsequent activity categories. 
Second, the activity evolution uncertainty causes the model to capture the spurious correlation between features and activity categories, which leads to unexplainable model behavior based on visually similar observed videos. 
The data uncertainty also brings noise into the intermediate representations and serious error accumulation for the generative anticipation methods~\cite{srl}. 
In previous work, the data uncertainty is addressed in a multi-label classification framework~\cite{camporese2021knowledge} or estimated by prediction distributions sampling~\cite{abu2019uncertainty}. Another general solution is to resort to prior knowledge to intervene in the anticipation or auxiliary prediction tasks on verbs and nouns to reduce the uncertainty~\cite{furnari2018leveraging}. 
Unfortunately, without comprehensively exploring data uncertainty factors in model learning, the robustness and generality of the models can hardly be guaranteed. 
Accordingly, we propose an uncertainty-boosted generative video activity anticipation framework, which models the data uncertainty from the intrinsic temporal correlation of activity classes and the semantic relationship from the external commonsense knowledge base.

Specifically, in parallel with the anticipation outputs, our model produces an uncertainty value with a simple fully connected layer to indicate the credibility of the outputs. Large uncertainty values indicate more diverse but less reliable model outputs, while small uncertainty values indicate more determined and trustable model outputs.
We employ this uncertainty value as a temperature parameter of a softmax activation function, to adjust the smoothness of the produced probability distribution of the target activity categories. 
Considering the activity evolution characteristics, we construct a target activity category label space to guarantee the output adjustment, which represents the probability of the target activity category and the set of activity classes that may co-occur with the target activity. The set of co-occurred classes is the union of the categories from the temporal activity cooccurrence statistics in a training dataset and the correlated activity class from the ConceptNet~\cite{speer2017conceptnet5}. 
It provides a way to quantitatively measure activity evolution uncertainty and reveals the temporal evolution pattern among a set of closely related visual concepts. 
Our target label space approximates the target activity probability distribution more comprehensively than the one-hot label paradigm. On the other hand, it reduces the spurious correlation globally existing among the whole label set in the multi-label paradigm, gaining more flexibility and expressiveness in modeling context dependency.

Another major issue is the estimation mechanism of uncertainty value. Quantifying the uncertainty into an absolute value with a universal rule applicable to all samples appears to be challenging and untractable. Instead, given a set of video instances, it is more reasonable to determine the relative values of their uncertainty, which can be calculated from the sample-wise and temporal perspectives. 
The former is modeled according to the relative anticipation difficulty among training samples. 
It first produces a weighted combination of the anticipated features from a set of samples with respect to their normalized uncertainty values. The model is then forced to predict the target activity categories of the samples in the mixed feature. During the training, the model has to borrow more information from those hard examples to ensure their anticipation accuracy, so that the weights (the relative uncertainty values) of these samples will be encouraged to be larger than others. 
For the temporal uncertainty modeling, it is assumed that the uncertainty value will be gradually reduced as the observed time length $\tau_o$ increases and the anticipation time length $\tau_a$ decreases. Accordingly, we design a temporal uncertainty ranking loss function to learn the relative order of the uncertainty values between samples with different anticipation time lengths for the same target activity class. This ensures the uncertainty of the output to be estimated comprehensively and regularly.

We apply our method to multiple backbones on EPIC-KITCHENS-55~\cite{damen2018scaling}, EPIC-KITCHENS-100~\cite{EPIC-100}, EGATE Gaze+\cite{li2018eye}, MECCANO~\cite{ragusa2023meccano} and 50 Salads~\cite{stein2013combining}. Experimental results demonstrate that our method achieves more comprehensive uncertainty modeling ability, better robustness/explainability, and remarkable anticipation performance. 
The proposed strategy can also inspire a broad spectrum of video comprehension tasks~\cite{wu2022memvit,babaeizadeh2018stochastic,rudenko2020human,wang2021self}. 
For instance, the way of modeling data uncertainty facilitates full dataset exploitation for long-term video understanding~\cite{wu2022memvit}, forecasting~\cite{babaeizadeh2018stochastic} and trajectory prediction~\cite{rudenko2020human} tasks. It can also inspire the design of novel proxy tasks~\cite{wang2021self} for self-supervised video representation learning towards video foundation model construction.

The contribution is highlighted as follows:
\begin{itemize}
	\item We propose an uncertainty-boosted activity anticipation framework to enhance the output robustness by comprehensively exploring the data uncertainty in the video content and activity evolution. 
	\item By incorporating activity evolution from temporal class correlation and semantic relationship, the output uncertainty, measured in sample-wise and temporal-wise relative manner, fully reflects the uncertainty across samples and categories, and is used to modulate the predicted activity distribution and gain improved model generalizability.
	\item Experiments on multiple benchmarks demonstrate the effectiveness of our framework in terms of improved accuracy and more robustness/interpretability over existing models, especially when dealing with highly uncertain samples and long-tailed activity categories.  
\end{itemize}

\section{Related Work}
\label{sec:related-work}

\subsection{Activity Anticipation} 
\label{subsec:anticity_anticipation}

Video activity anticipation has achieved rapid development~\cite{wu2017anticipating,fan2018forecasting,rhinehart2017first,zhang2017deep,zhang2020egocentric,felsen2017will,zhao2020diverse,zhang20anegocentric,nawhal2022rethinking, zhang2024object, wang2023memory, girase2023latency, thakur2024leveraging}.These methods are roughly divided into generative and non-generative families~\cite{hutchinson2021video}.

\textbf{Generative Anticipation Methods.}
This paradigm first anticipates future features and then performs activity classification. 
Furnari {\it et al.}~\cite{rulstm} introduce the RULSTM, which processes RGB, optical flow and object-based features using two LSTMs and a modality attention mechanism to anticipate future activities. 
Qi {\it et al.}~\cite{srl} propose a self-regulated learning framework for activity anticipation. 
Fernando {\it et al.}~\cite{fernando2021anticipating} propose to correlate past features with the future using three similarity measures. 
Girdhar{\it et al.}~\cite{activipative_transformer} propose an end-to-end attention-based video modeling architecture that attends to the previously observed video to anticipate future activities. 
Zhong {\it et al.}~\cite{zhong2023diffant} introduce a diffusion model-based approach for long-term action anticipation. 
Some recent works~\cite{yu2023merlin, zhao2023antgpt, kim2023lalm} utilize Multimodal Large-Language Models to assist in predicting future activities.

\textbf{Non-generative Anticipation Methods.}
This paradigm often directly generates the anticipation results based on the observed video. 
Aliakbarian {\it et al.}~\cite{sadegh2017encouraging} propose a multi-stage LSTM architecture that leverages context-aware and activity-aware features and develop a loss function that encourages the model to predict the correct class as early as possible. 
Qi {\it et al.}~\cite{qi2017predicting} propose a spatial-temporal And-Or graph to represent events, and an early parsing method using temporal grammar to anticipate the next activity.
Farha {\it et al.}~\cite{abu2019uncertainty} propose to sample multiple times to estimate the predicted distributions at the test stage. 
Ke {\it et al.}~\cite{ke2019time} propose to explicitly condition the anticipation on time, which is shown to be efficient and effective for long-term activity anticipation. 
Mahmud {\it et al.}~\cite{mahmud2017joint} propose a hybrid Siamese network for jointly predicting the label and the starting time of future unobserved activity.

\subsection{Uncertainty Learning} 
\label{subsec:ul}

As discussed in~\cite{der2009aleatory, kendall2017uncertainties}, there are two main types of uncertainty, {\it i.e.}, aleatory and epistemic uncertainty, also known as data and model uncertainty. 
In recent years, a major effort has been dedicated to quantifying data and model uncertainty for various tasks in natural language processing (NLP)~\cite{kendall2017uncertainties, xiao2019quantifying, gal2016theoretically, zhu2017deep} and computer vision~\cite{kendall2017uncertainties, kendall2015bayesian, isobe2017deep,choi2019gaussian,yu2019robust,chang2020data,zhang2021relative,chen2020monopair,zhou2021model,feng2018towards}.

In detail, Xiao {\it et al.}~\cite{xiao2019quantifying} study the benefits of characterizing model and data uncertainty, which show that explicitly modeling uncertainty is useful for measuring output confidence and enhancing performances in various NLP tasks. 
Kendall {\it et al.}~\cite{kendall2017uncertainties} study the benefits of modeling epistemic and aleatoric uncertainty in Bayesian deep learning models for per-pixel semantic segmentation and depth regression tasks, which achieves promising results on multiple benchmarks. 
Zhang {\it et al.}~\cite{zhang2021relative} regard uncertainty as a relative concept and propose a Relative Uncertainty Learning method, which performs well on both real-world and synthetic noisy facial expression recognition datasets.

\begin{figure*}[t]
	\centering
	\includegraphics[scale=0.67]{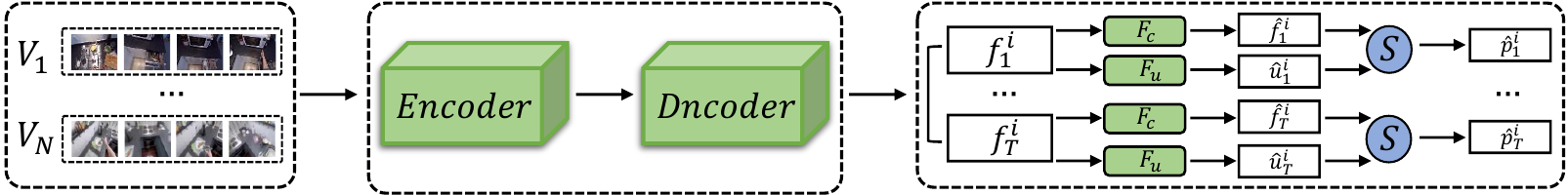}
	\caption{The uncertainty-boosted activity anticipation framework. We use $F_c$ and $F_u$ to produce the probability distribution of the anticipation result and the uncertainty vector. Then we use $\hat{u}^i_t$, the mean of the uncertainty vector, to adjust the smoothness of the distribution ($S$) and obtain $\hat{p}^i_t$.}
    \vspace{-4mm}
	\label{fig_uncertianty_learning_framework}
\end{figure*}

The relative uncertainty learning method on images~\cite{zhang2021relative} is similar to ours. Considering the characteristics of the video anticipation task, our relative uncertainty learning strategy is constructed from two complementary perspectives, namely, sample-wise and temporal. 
In the sample-wise relative uncertainty learning, we infer the target activity labels based on the video evolution characteristic, which guarantees more comprehensive uncertainty modeling compared to the one-hot label representation used by traditional methods. The new proposed temporal relative uncertainty learning assumes that the uncertainty values will be gradually reduced as the observed time length increases and the anticipation time length decreases, which further enhances the rationality of the relative uncertainty values.

\subsection{Explanability in Video Analytics} 
\label{subsec:eai}

The development of theories, frameworks, and tools to explainable artificial intelligence has become an active research field~\cite{confalonieri2021historical, longo2020explainable} recently. In video understanding, the explainability/interpretability has been investigated and improved from data explanation~\cite{zhi2021mgsampler, szymanowicz2021x}, stateful procedure~\cite{zhuo2019explainable} and middle-level concept relation~\cite{qi2020modeling,qi2020towards} aspects. 
In this paper, we address the interpretability issue by uncertainty modeling, {\it i.e.}, to quantify the uncertainty in the anticipation process and output from complementary aspects, which facilitates more regulated model learning compared to the end-to-end data fitting paradigms. We further prove that quantifying the uncertainty can enhance both the model interpretability and performance, rather than seeking a compromise between the two~\cite{longo2020explainable}.

\begin{figure}[t]
    \centering
    \includegraphics[scale=0.42]{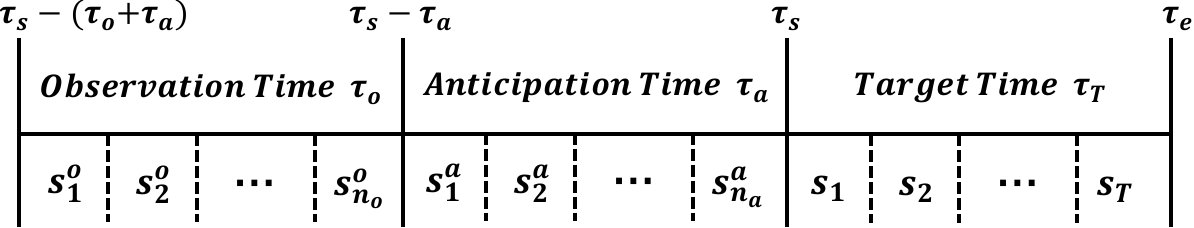}
    \caption{The setting of the video activity anticipation task.}
    \vspace{-4mm}
    \label{fig_taskintro}
\end{figure}

\section{Method}
\label{sec:method} 

As shown in~\Cref{fig_taskintro}, given an observed video of length $\tau_o$ that starts at $\tau_s-(\tau_o + \tau_a)$ and ends at $\tau_s-\tau_a$,  the video activity anticipation task aims to predict what will happen after $\tau_a$ by anticipating the activity categories of a video starting at $\tau_s$ and ending at $\tau_e$~\cite{damen2018scaling}. 
Here, $\tau_o$, $\tau_a$ and $\tau_T$ represent the observation time length, the anticipation time length, and the target activity time length, respectively. 
Following the widely adopted protocol in~\cite{rulstm, srl, xu2022learning}, 
we evenly divide the video into snippets every $\delta$ seconds. 
Then we obtain $n_o$ snippets in the observed time interval $[\tau_s-(\tau_o + \tau_a), \tau_s-\tau_a]$, $n_a$ snippets in the anticipation time interval $[\tau_s-\tau_a, \tau_s]$ and $T$ snippets in the target activity time interval $[\tau_s, \tau_e]$, represented by $S_o=\left\{{s^o_1, s^o_2, ..., s^o_{n_o}}\right\}$, $S_a=\left\{{s^a_1, s^a_2, ..., s^a_{n_a}}\right\}$ and $S_T=\left\{{s_1, s_2, ..., s_T}\right\}$, respectively.

\subsection{Generative Video Activity Anticipation Framework}
\label{subsec_generativeframework}

The generative activity anticipation framework first generates future feature representations and then performs target activity anticipation. 
Given observed videos $\left\{V_1, ..., V_N \right\}$, we first utilize an encoder-decoder to model the observed video content and produce future feature representations $\left\{f^i_1, ..., f^i_T\right\}$ for each observed video $V_i$. 
Then, we employ a fully connected layer with a softmax activation function to perform target activity anticipation, and obtain the results $\left\{p^i_1, ..., p^i_T\right\}$, where $p^i_t\in{\mathbb{R}^{C}}$ and $C$ is the number of the target activity categories. 
It should be noticed that existing solutions differ significantly in the design of the encoder-decoder. 
For example, the classical method RULSTM~\cite{rulstm} is based on the Rolling-Unrolling LSTMs, while the more recent method DCR~\cite{xu2022learning} utilizes a transformer-based module. Either of them is applicable in our framework.

\subsection{Uncertainty-boosted Video Activity Anticipation}
\label{subsec_ulanticipation}

\subsubsection{Overview}
\label{subsec_overview}

For activity anticipation, we aim to produce an uncertainty value from the probability distribution of the potential target activity categories, which indicates the reliability of the model outputs. As shown in~\Cref{fig_uncertianty_learning_framework}, we initially utilize an encoder-decoder module to generate future feature representations $\left\{f^i_1, ..., f^i_T\right\}$ for each observed video $V_i$ at all anticipation times. This encoder-decoder module can be LSTM-based, transformer-based, or of other types. Then, we utilize two parallel fully connected layers to perform uncertainty-boosted target activity anticipation. Specifically, $F_c$ is utilized to produce the probability distribution related to the target activity, and $F_u$ is employed to generate the uncertainty vector $u^i_t\in{\mathbb{R}^{C}}$ of the anticipation output. 
In line with~\cite{chang2020data, zhang2021relative}, we consider the uncertainties in both video data and activity evolution within the uncertainty vector and take the mean of $u^i_t$ as the approximated uncertainty value $\hat{u}^i_t$ for sample $i$ at time $t$. The mean operation serves to mitigate the influence of noise and outliers within the uncertainty vector. 
The above strategy produces an all-embracing and robust uncertainty estimate result, benefiting further optimization of the predicted probability distribution. 	
We calculate $\hat{p}^i_t$ by adjusting the smoothness of the target activity probability distribution. 
Presenting uncertainty in the form of scalar values simplifies the interpretation and utilization of uncertainty estimation results, allowing for comparison across different models and datasets.

\begin{figure}[t]
	\centering
	\includegraphics[scale=0.46]{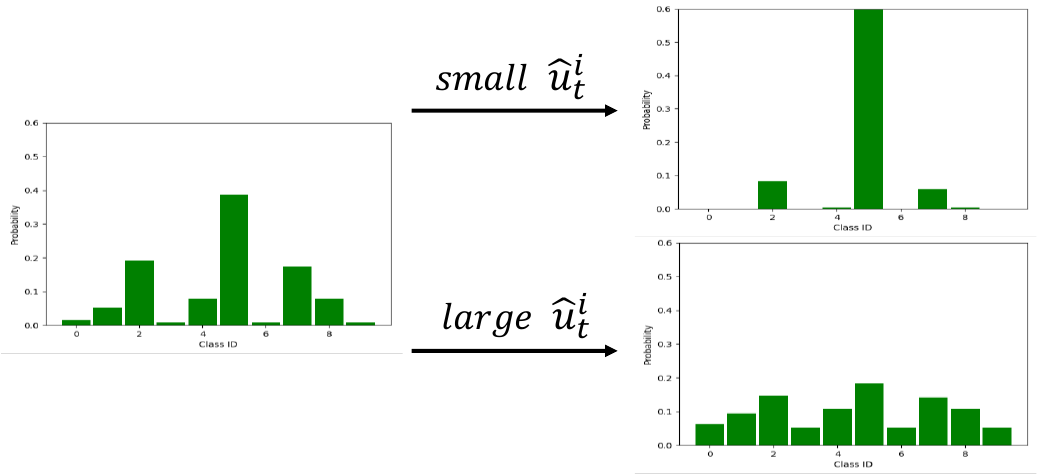}
	\caption{Given a probability distribution of the anticipation result, we visualize the effects of the distribution adjustment strategy.}
     \vspace{-4mm}
	\label{fig_uncertainty_scale}
\end{figure}

\subsubsection{Distribution Adjustment}
\label{subsec_disadjustment}

A small uncertainty value indicates the anticipation result is trustworthy. In this case, the model output should focus on a few target activity categories, {\it i.e.}, the probability distribution of the potential target activity categories tends to be sharp. Instead, a large uncertainty value means the result is not reliable enough. The model should produce multiple possible target activity categories and the probability distribution should be flat. 
Our distribution adjustment strategy treats the mean uncertainty value $\hat{u}_t^i$ as a temperature parameter of the softmax function to adjust the smoothness of the target activity probability distribution and produce $\hat{p}^i_t = \left\{\hat{p}^i_{t,1}, ..., \hat{p}^i_{t,C}\right\}\in{\mathbb{R}^{C}}$ through, 
\begin{equation}
	\hat{p}^i_{t, j} = \frac{\exp(\hat{f}^i_{t,j}/\hat{u}^i_t)}{\sum^{C}_{r=1}{\exp(\hat{f}^i_{t,r}/\hat{u}^i_t)}},
\end{equation}
where $\hat{f}^i_t = \left\{\hat{f}^i_{t,1}, ..., \hat{f}^i_{t,C}\right\}$. 
The adjusted distribution gives more calibrated anticipation results. As shown in~\Cref{fig_uncertainty_scale}, the probability distribution will be sharp with small uncertainty values. Instead, it will be flat. 

To optimize the above probability distribution and guarantee this adjustment effect, we construct a target activity label space to represent the probability of the target activity category and the set of activity classes that may co-occur with the target activity. 
First, we obtain the uncertainty matrices representing the possibility between any pair of activity classes that may co-occur in parallel in the future based on the same antecedent activity, which is computed from two complementary sources. 
One is the internal temporal co-occurrence information of activity classes contained in all videos of the dataset. 
Especially, for any two activity classes, we count the number of common antecedent class instances they share across all videos. For example, if the dataset contains activity sequences `open fridge, take milk' and `open fridge, close fridge', then the number of common antecedent activity class instances for class pair `take milk' and `close fridge' increases by 1. By analyzing the entire dataset, we get the internal uncertainty matrix $\boldsymbol{Rl_{i}} \in{\mathbb{R}^{C \times C}}$ of activity classes. 
Each entry $(i,j)$ in this matrix represents the number of instances that the categories $c_i$ and $c_j$ evolve from the same antecedent activity class in the dataset.

The other is the activity class relationships from the external commonsense knowledge graph ConceptNet. 
Activity categories with meaningful edges in ConceptNet are semantically dependent. For example, `HasSubevent' means the activity categories have an inclusion relationship. This is conducive to effectively expanding the target activity category.  
Specifically, for commonly used datasets, its activity class is a (verb, noun) pair, so we can acquire $C_v$ unique verb classes and $C_n$ unique noun classes. 
For any two verb classes, we calculate the number of connected paths with only one intermediate node under the selected relationships in ConceptNet, and get the external uncertainty matrix $\boldsymbol{Rl^v_{e}} \in{\mathbb{R}^{C_v \times C_v}}$ of verb classes. Similarly, we obtain the external uncertainty matrix $\boldsymbol{Rl^n_{e}} \in{\mathbb{R}^{C_n \times C_n}}$ of noun classes. 
The external uncertainty score of an activity class is captured by simply adding the external uncertainty of its verb class and noun class. Finally, we acquire the whole external uncertainty matrix $\boldsymbol{Rl_{e}} \in{\mathbb{R}^{C \times C}}$ of activity classes. 

Next, we generate an activity class set $\mathcal{A}^c$ that may co-occur in parallel with the target activity class $c$. 
We obtain the internal uncertainty vector $Rl^c_{i}$ and the external uncertainty vector $Rl^c_{e}$ from the $c$-th row of $\boldsymbol{Rl_{i}}$ and $\boldsymbol{Rl_{e}}$. 
Then, we obtain $\mathcal{A}^c$ by simply merging $Rl^c_{i}$ and $Rl^c_{e}$ according to their value and eliminating classes with zero value. 

Finally, given sample $V_i$ with target activity category $c$, the ideal target activities may be the given category $c$ or categories contained in $\mathcal{A}^c$. Only categories in $\mathcal{A}^c$ have a low probability of occurrence. We construct the target activity category label representation $\overline{p}^i_t = \left\{\overline{p}^i_{t,1}, ..., \overline{p}^i_{t,C}\right\}\in{\mathbb{R}^{C}}$, 
\begin{equation}
	\overline{p}^i_{t,j}=
	\begin{cases}
		1-\alpha \quad &j=c,\\
		\frac{\alpha}{|\mathcal{A}_c|} \quad &j \in \mathcal{A}_c,\\
		0 \quad &others,
	\end{cases} 
	\label{eqa_target_label}
\end{equation}
where $\alpha$ is the total weight of activity classes in $\mathcal{A}^c$. $|\mathcal{A}_c|$ is the number of categories in $\mathcal{A}_c$. 
We do not use the values in $Rl^c_{i}$ and $Rl^c_{e}$ as $\overline{p}^i_{t,j}$ for categories in $\mathcal{A}^c$, since it may cause over-fitting and reduce the model generalization ability. 

The loss function of the anticipation model is as follows, 
\begin{equation}
	L_{c} =-\sum_i \Big\{(1-\alpha)\log{\hat{p}^i_{t,c}} + \sum_{c^{'}\in\mathcal{A}_c}{\frac{\alpha}{|\mathcal{A}_c|}\log{\hat{p}^i_{t,c^{'}}}}\Big\},
\end{equation}
It guarantees the anticipation model to learn both the deterministic target activity class $c$ and activity classes in $\mathcal{A}_c$ that may co-occur with the target activity.

\begin{figure*}[t]
	\centering
	\includegraphics[scale=0.7]{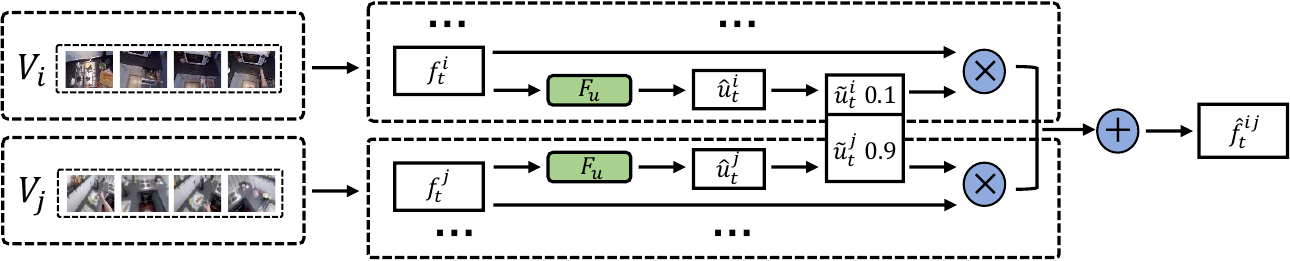}
	\caption{The sample-wise relative uncertainty learning pipeline. We obtain $\hat{f}^{ij}_t$ by mixing the anticipated features with their relative uncertainty values.}
  \vspace{-4mm}
	\label{fig_uncertianty_mixup}
\end{figure*}

The constructed labels, derived through statistical analysis of the task dataset and enhanced with an external knowledge base, truly reflect the correlation between visual concepts and activity labels in the video content. Consequently, the anticipation model will consider a broader range of potential future activities, allowing it to accurately grasp activity evolution patterns and avoid the spurious correlations between feature and prediction output caused by dataset bias compared to the traditional one-hot label strategy. 
In contrast to multi-label learning techniques, our strategy does not treat multiple labels indiscriminately. Instead, it empowers the model to unearth multiple potential activities while ensuring the primary accuracy for the given target activity. 
Compared with the label smoothing strategy in the multi-label paradigm, our approach refrains from assigning uniformly low probability values to all classes. Instead, it identifies the class set to which probability values should be assigned based on the association between activity classes. 
Our soft constraint prevents the model from over-fitting into one single category and diminishes the impact of potential label noise, enhancing overall generalization ability. 
In addition, commonly used activity anticipation datasets suffer from serious long-tail distribution problems. 
As shown in Equation (2), our strategy associates majority classes with minority classes, and augments datasets to balance the frequency of activity categories. This prevents the model from over-training on extremely frequent activities and mitigates the long-tail distribution issue from the data sampling perspective. 
Our strategy also provides solutions for the challenging problem of quantitatively analyzing uncertainty, particularly on the complex video content, which will contribute to explainable video understanding.

\subsubsection{Relative Uncertainty Learning}
\label{subsec_relativeul}

It is straightforward to represent and learn the uncertainty in absolute value~\cite{chang2020data,wang2020suppressing}. However, deep neural networks have strong learning and memory ability, which tends to remember hard examples~\cite{chang2020data, wang2020suppressing, zhang2021relative}. 
In fact, without comparison, it is difficult to judge whether the absolute uncertainty value is reasonable or accurate. 
Given a set of samples, one can easily estimate the relative order of their uncertainty values. 
To this end, we model the relative uncertainty from the sample-wise and temporal perspectives.

\textbf{Sample-wise Relative Uncertainty.} 
Given a set of samples $V=\left\{V_1, ..., V_N \right\}$, the model generates feature representations $f_t=\left\{f^1_t, ..., f^N_t\right\}$ and uncertainty values $U_t = \left\{\hat{u}^1_t, ..., \hat{u}^N_t\right\}$ for each anticipation time $t$. 
Then we can get the relative uncertainty values $\widetilde{U}_t=\left\{\widetilde{u}^1_t, ..., \widetilde{u}^N_t\right\}$ by dividing $U_t$ by the sum of all the uncertainty values. 
We take the weighted sum of $f_t$ over $\widetilde{U}_t$ and get the mixed feature $\hat{f}_t$. 
The model is forced to predict the target activity categories contained in samples $V$ based on $\hat{f}_t$. 
As the training progresses, the model will predict the target activity categories of easy samples, but still cannot predict the activity labels of the hard samples well, {\it i.e.}, they are assigned with larger uncertainty values. This indicates a relatively large classification loss on the prediction of hard samples, which in turn forces the mixed feature to contain more information from the hard samples to achieve a smaller activity anticipation loss value, according to the weight combination using the uncertainty values to produce the mixed sample.  
This strategy ensures that the anticipation model appropriately assigns larger uncertainty values on hard training samples by comparing them to others.

Without loss of generality, we detail the strategy in the pair-wise case. 
As shown in~\Cref{fig_uncertianty_mixup}, given two training samples $V_i$ and $V_j$ with different target activity classes in the same mini-batch, we obtain feature representations $f^i_t$ and $f^j_t$ and the corresponding uncertainty values $\hat{u}^i_t$ and $\hat{u}^j_t$. Then, we get the relative uncertainty values and the mixed future feature representation $\hat{f}^{ij}_t$ through, 
\begin{equation}
	\widetilde{u}^i_t = \frac{\hat{u}^i_t}{\hat{u}^i_t + \hat{u}^j_t}, \widetilde{u}^j_t = \frac{\hat{u}^j_t}{\hat{u}^i_t + \hat{u}^j_t},
\end{equation}
\begin{equation}
	\hat{f}^{ij}_t = \widetilde{u}^i_t f^i_t + \widetilde{u}^j_t f^j_t.
	\label{eqa_srul}
\end{equation}
Finally, we get the anticipation result $\hat{p}^{ij}_t\in{\mathbb{R}^{C}}$ through $\hat{f}^{ij}_t$. 
Since the target activity category label should also be a mixture of target activity category labels of the two samples, we construct the following target activity category label $\overline{p}^{i,j}_t = \left\{\overline{p}^{i,j}_{t,1}, ..., \overline{p}^{i,j}_{t,C}\right\}\in{\mathbb{R}^{C}}$ according to~\Cref{subsec_disadjustment}, 
\begin{equation}
	\overline{p}^{i,j}_{t,k}=
	\begin{cases}
		\frac{1-\alpha}{2} \quad &k=i,j,\\
		\frac{\alpha}{|\mathcal{A}_{c_i,c_j}|} \quad &k \in \mathcal{A}_{c_i,c_j},\\
		0 \quad &others,
	\end{cases} 
\end{equation}
where $c_i$ and $c_j$ represent the target activity category of sample $V_i$ and $V_j$, respectively. $\alpha$ is the total weight of the activity classes contained in $\mathcal{A}_{c_i,c_j}$, which is the activity class set that may co-occur with the target activity classes of samples $V_i$ and $V_j$. Correspondingly, we propose the following sample-wise relative uncertainty loss function, 
\begin{equation}
	\begin{split}
		L_{srul} = -\sum_{k}\frac{1}{B}\sum_{i,j}^{B}
		\Big\{&\frac{1-\alpha}{2}\left( \log{\hat{p}^{ij}_{t,c_i}}+\log{\hat{p}^{ij}_{t,c_j}} \right)\\ &+\sum_{c^{'}\in\mathcal{A}_{c_i,c_j}}\frac{\alpha}{|\mathcal{A}_{c_i,c_j}|}\log{\hat{p}^{ij}_{t,c^{'}}}
		\Big\},
	\end{split}
	\label{eq_sample_relative_loss}
\end{equation}
where $B$ is the mini-batch size.

When there are multiple samples in $V$, the effectiveness of the sample-wise relative uncertainty learning strategy will be reduced. 
Since the mini-batch is randomly generated, we can not ensure the diversity of samples in the mini-batch. 
If these samples have similar anticipation difficulty, the generated relative uncertainty values between them will not vary much. 
Similar relative uncertainty values will cause the mixed feature to contain a similar amount of information from each sample.
This can not force the anticipation model to capture useful features for hard samples and hinders the validity of the~\Cref{eq_sample_relative_loss}. 

Similar to~\cite{zhang2021relative}, we construct mixed samples according to the relative uncertainty values. 
The difference is that we introduce more reasonable target activity labels for the mixed samples based on the activity evolution uncertainty, instead of using one-hot labels as in~\cite{zhang2021relative}. \Cref{eq_sample_relative_loss} enables the model to predict the two target activity categories equally based on the mixed features and ensures that the model can also predict the activities that have a co-occurrence relationship with the target activities. This guarantees more comprehensive uncertainty modeling results.

\begin{figure}[t]
	\centering
	\includegraphics[scale=0.4]{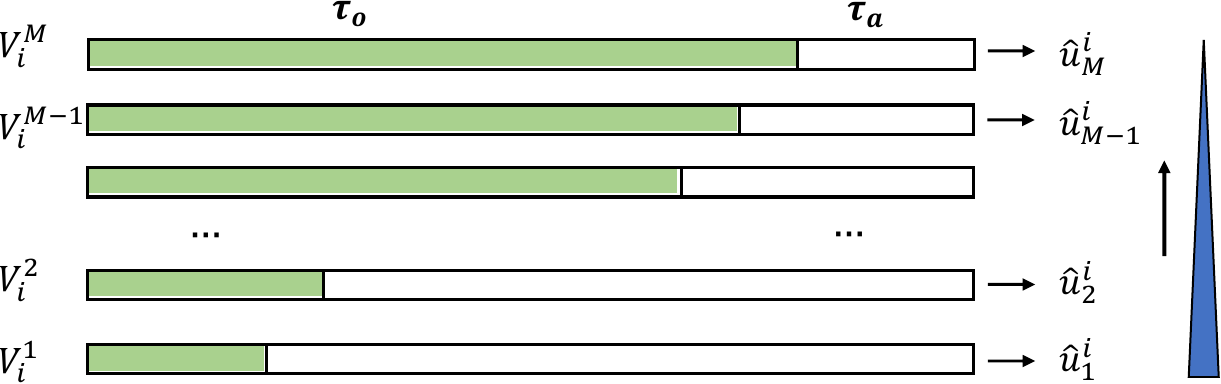}
	\caption{The examples of the temporal relative uncertainty learning. If we gradually expand the observed time length $\tau_o$ and reduce the anticipation time length $\tau_a$, the uncertainty values should gradually decrease.}
  \vspace{-4mm}
	\label{fig_uncertainty_listmle}
\end{figure}

\textbf{Temporal Relative Uncertainty.} 
Given the same target activity, if we gradually enlarge the observation time length $\tau_o$ and correspondingly reduce the anticipation time length $\tau_a$, the anticipation difficulty will gradually decrease, and the outputs will become more reliable, {\it i.e.}, the corresponding uncertainty values will gradually decrease. 
We propose to learn the relative order of uncertainty values according to the listwise ranking paradigm~\cite{xia2008listwise}. 

Given an observed video $V_i$, we first prepare $M$ training samples $\left\{V_i^1, V_i^2, ..., V_i^M \right\}$ with the same target activity category. 
As shown in~\Cref{fig_uncertainty_listmle}, from sample $V_i^1$ to $V_i^M$, their observation video length $\tau_o$ are gradually expanded and the corresponding anticipation time length $\tau_a$ are gradually reduced. Then, the anticipation model will generate uncertainty values $U^i=\left\{\hat{u}_{1}^i, \hat{u}_{2}^i, ..., \hat{u}_{M}^i \right\}$. 
We rank these uncertainty values from the largest to the smallest as $\hat{\pi}_i=\left\{R^1_k, R^2_M, ..., R^M_2 \right\}$. $R^l_j$ means that the sample $V^l_i$ has the $j$-th largest uncertainty value among all samples. 
Ideally, the uncertainty values will gradually decrease and the ranking result will be $\pi_i =\left\{R^1_1, R^2_2, ..., R^M_M \right\}$. 
Based on this, we introduce the following permutation probability distribution~\cite{xia2008listwise}, which means the probability of obtaining sequence $\pi_i$ given the uncertainty values $U^i$, 
\begin{equation}
	P(\pi_i|U^i)=\prod_{j=1}^M \frac{\phi(U^i_{\pi_i^{-1}(j)})}{\sum^M_{m=j}\phi(U^i_{\pi_i^{-1}(m)})},
\end{equation}
where $\phi$ is the identity mapping function. 
$\pi_i^{-1}(j)$ represents the sample at the $j$-th position in sequence $\pi_i$, {\it i.e.}, sample $V^j_i$. 
$U^i_{\pi_i^{-1}(j)}$ represents its uncertainty value $\hat{u}_{j}^i$ from $U^i$. 

A well-trained anticipation model should produce a $\hat{\pi}_i$ that is very close to or even the same as $\pi_i$. Accordingly, $P(\pi_i|U^i)$ will go as large as possible. 
If we optimize the negative $\log$ of $P(\pi_i|U^i)$, we can ensure that the model can produce rational uncertainty ranking results. 
Therefore, we propose the following temporal uncertainty ranking loss function to learn the relative order of the uncertainty values, 
\begin{equation}
	L_{trul} =-\sum_i \log P(\pi_i|U^i). 
\end{equation}

\subsubsection{Training Objective}
\label{subsec_trainingobjective}

The whole learning objective function of our framework is, 
\begin{equation}
	\begin{split}
		L = L_{srul} + \beta L_{trul} + \gamma L_{wd},
	\end{split}
\end{equation}
where $\beta$ and $\gamma$ are the corresponding loss weights. Since we employ the sample-wise relative uncertainty learning strategy, the $L_c$ is fused into $L_{srul}$. $L_{wd}$ is the sum of the squares of all uncertainty values, which can be seen as a regularization on the uncertainty representation. This loss is used to keep the uncertainty values stable.

\subsubsection{Discussions}

Compared with existing generative activity anticipation methods, our framework only requires an additional fully connected layer to generate uncertainty values indicating the reliability of the output. It can be inserted into most existing methods with a small cost. 
They can be endowed with better generalization ability when dealing with samples with high uncertainty, and activity categories with high uncertainty or long-tailed distributions. 

Furthermore, our method enlightens a wide range of video comprehension tasks. 
First, through quantitative analysis of the data uncertainty, we have a deep insight into different video activity datasets. This is beneficial for the data usage of video understanding tasks. Selecting data modality with less uncertainty is more conducive to capturing temporal association relationships. 
Second, the uncertainty is common in other sequence modeling or prediction tasks such as video forecasting~\cite{babaeizadeh2018stochastic} and trajectory prediction~\cite{rudenko2020human}. Our distribution adjustment and relative uncertainty learning strategies can be applied to these tasks as well. 
Third, mining category co-occurrence relationships and constructing precise labels are applicable to long-term video understanding~\cite{wu2022memvit}. Compared with one-hot labels and label smoothing, the constructed labels are more conducive to producing effective video representation. 
Finally, the idea of the temporal relative uncertainty learning strategy can be used as a new proxy task for self-supervised video representation learning~\cite{wang2021self}. 
This also paves the way towards the construction of video foundation models and better model adaptation on various down-stream applications.

\section{Experiments}
\label{sec:experiments}

\subsection{Setups}
\label{subsec_setup}

\subsubsection{Datasets} 
\textbf{EPIC-KITCHENS-55 (EK55)}~\cite{damen2018scaling} is a first-person cooking dataset captured by 32 subjects in 32 kitchens. 
It includes 125 verb classes and 352 noun classes. 
We consider all the unique (verb, noun) class pairs in the training set, and obtain 2513 activity classes.  
Following~\cite{rulstm}, we randomly split the train set into train and validation sets. 
The test set splits into sets with seen kitchens~(\textbf{S1}) and unseen kitchens~(\textbf{S2}). 

\textbf{EPIC-KITCHENS-100  (EK100)}~\cite{EPIC-100} is a substantial extension of EK55 dataset, which contains 89,977 segments of activity. It has 97 verb classes, 300 noun classes, and 4053 activity classes. The validation/test splits contain two subsets, Unseen Participants and Tail Classes. 

\textbf{EGTEA Gaze+}~\cite{li2018eye} records the first-person meal preparation activities, which includes 10325 instances with 106 activity classes. It provides three different train/test splits, and we report the average performance across all splits. 

\textbf{MECCANO}~\cite{ragusa2023meccano} is a multimodal dataset related to human behavior understanding in industrial-like settings (e.g., factories, building sites, mechanical workshops). 
It offers 20 sequences with 299,376 annotated frames from 20 different participants. It has 12 verbs, 20 nouns, and 61 unique actions. 
We employ the standard train-val-test split.

\textbf{50 Salads}~\cite{stein2013combining} encompasses 50 videos documenting salad preparation tasks executed by 25 distinct actors. These videos encompass 17 fine-grained activity categories. We discern 7 unique verb classes and 14 noun classes from the provided activity categories. Following~\cite{stein2013combining}, we adopt a five-fold cross-validation strategy for assessment purposes.

\begin{table}[t]
	\centering
	\small
	\renewcommand{\arraystretch}{1}
	\setlength{\tabcolsep}{0.8mm}
	\caption{Parameter values of $\alpha, \beta, \gamma$ in our framework.}
        \vspace{-2mm}
	\begin{tabular}{c c c c}
		\toprule
		& Ub-Baseline & Ub-RULSTM & Ub-DCR \\
		\midrule
		EK-55 & 0.4/0.005/5e-6 & 0.4/0.05/1e-4 & 0.4/5/1e-3 \\
		EK-100 & / & 0.2/0.01/1e-3 & 0.1/1/1e-4 \\
		EGTEA Gaze+ & / & 0.1/0.5/1e-4 & 0.1/20/1e-2 \\
		MECCANO & / & 0.4/0.1/1e-3 & 0.4/1e-3/1e-3 \\
		\bottomrule
	\end{tabular}
        \vspace{-4mm}
	\label{table_params}
\end{table}

\subsubsection{Metrics}
For the EK55 validation set, we use top-5 accuracy and mean top-5 recall as measurements. The mean top-5 recall is averaged over the provided list of many-shot verbs, nouns and activities. 
For the EK55 test set, we use top-1 and top-5 accuracy as measurements. 
For EK100, we report the top-5 recall. For EGTEA Gaze+ and MECCANO, we report the top-5 accuracy.
For the 50 Salads, we use mean accuracy over classes for performance comparison.

\subsubsection{Different Backbones}

\textbf{Baseline.} The encoder-decoder in~\Cref{fig_uncertianty_learning_framework} is chosen to be a GRU-GRU block. We set the observed video length as 1.5s and choose $\delta$ as 0.25s. 
Two parallel fully connected layers are used to produce the probability distribution of the target activity categories and the corresponding uncertainty value. 
The anticipation video snippets $n_a$ are set to 8. The model will predict the target activity classes at future timestamps 0.25s, 0.5s, 0.75s, 1s, 1.25s, 1.5s, 1.75s and 2s. 

\textbf{RULSTM.} Its encoder-decoder module is the Rolling-Unrolling LSTMs. For more details, please refer to~\cite{rulstm}. 

\textbf{DCR.} It employs transformer-based and rulstm-based encoder-decoder modules. To ensure the diversity of backbones, we choose the transformer-based DCR backbone. For more details, please refer to~\cite{xu2022learning}. 

\textbf{ActionBanks.} The encoder-decoder module is the Temporal Aggregation Block. More details are shown in~\cite{sener2020temporal}.

\textbf{FUTR.} The encoder-decoder module is built on self-attention and cross-attention. More details are shown in~\cite{gong2022future}.

In the experiments, we apply our model on top of the above-mentioned backbones, and the uncertainty boosted versions are denoted as Ub-Baseline, Ub-RULSTM, Ub-DCR, Ub-ActionBanks and Ub-FUTR respectively.

\subsubsection{Implementation Details}

For a fair comparison, we adopt the widely used multi-modality features. 
Concretely, for EK55 and EK100, we use the RGB feature from TSN~\cite{wang2016temporal}, the Flow feature from TSN, and the OBJ feature from FRCNN~\cite{girshick2015fast}, which are provided by~\cite{rulstm}. 
Besides, we utilize the RGB feature from TSM~\cite{lin2019tsm} provided by~\cite{xu2022learning}. 
We also use the RGB feature from irCSN-152~\cite{tran2019video} for EK55 provided by~\cite{activipative_transformer}. 
For EGTEA Gaze+, we use the RGB and Flow features from TSN provided by~\cite{rulstm}.
For MECCANO, we use the provided OBJ feature.

All experiments are implemented under the PyTorch framework. 
For the Baseline, we use an SGD optimizer with a mini-batch size of 128. 
The momentum and weight decay are set to 0.9 and 0.00005, respectively. The initial learning rate is set to 0.05. The number of total training epochs is 100. 
For other backbones, we follow the parameter settings of their original papers. 
For the parameters of our model, we record the optimal setting of $\alpha, \beta$ and $\gamma$ in~\Cref{table_params}. Besides, the $\alpha, \beta$ and $\gamma$ of Ub-FUTR are set to 0.2, 0.05, and 1e-4. All the parameters are determined via cross-validation.

\begin{table}[t]
	\centering
	\renewcommand{\arraystretch}{0.95}
	\setlength{\tabcolsep}{1.2mm}
	\caption{Apples-to-apples comparison on the EK100 validation set.}
        \vspace{-2mm}
	\begin{tabular}{c c c c c c}
		\toprule
		\multirow{2}{*}{Modality} & \multirow{2}{*}{Backbone} & \multirow{2}{*}{Model} & \multicolumn{3}{c}{Top-5 Recall @ 1s} \\
		\cmidrule{4-6}
		& & & Verb & Noun & Act \\
		\midrule
		\multirow{5}{*}{RGB} 
		& RULSTM & TSN & 27.5 & 29.0 & 13.3 \\
		& \bf Ub-RULSTM & TSN & \bf 28.3  & \bf 30.5  & \bf 14.3 \\
		\cmidrule{2-6}
		& DCR & TSN & 31.0 & 31.1 & 14.6 \\
		& \bf Ub-DCR & TSN & \bf 31.4  & \bf 31.8 & \bf 14.8 \\
		\cmidrule{2-6}
		& DCR & TSM & 32.6 & 32.7 & 16.1 \\
		& \bf Ub-DCR & TSM & \bf 32.8 & \bf 34.1 & \bf 16.2 \\
		\midrule
		\multirow{5}{*}{FLOW} 
		& RULSTM & TSN & 19.1 & 16.7 & 7.2 \\
		& \bf Ub-RULSTM & TSN & \bf 21.0 & \bf 16.8 & \bf 7.3 \\
		\cmidrule{2-6}
		& DCR & TSN & 25.9 & 17.6 & 8.4 \\
		& \bf Ub-DCR & TSN & \bf 26.8 & \bf 20.2 & \bf 8.7 \\
		\midrule
		\multirow{5}{*}{OBJ} 
		& RULSTM & FRCNN & 17.9 & 23.3 & 7.8 \\
		& \bf Ub-RULSTM & FRCNN & \bf 20.8  & \bf 24.5 & \bf 8.9  \\
		\cmidrule{2-6}
		& DCR & FRCNN & 22.2 & 24.2 & 9.7 \\
		& \bf Ub-DCR & FRCNN & \bf 23.1 & \bf 29.0 & \bf 10.9 \\
		\bottomrule
	\end{tabular}
        \vspace{-2mm}
	\label{table_epic100_modality}
\end{table}

\begin{table}[t]
	\centering
	\renewcommand{\arraystretch}{0.95}
	\setlength{\tabcolsep}{1.6mm}
	\caption{Apples-to-apples comparison on the EK55 validation set.}
        \vspace{-2mm}
	\begin{tabular}{c c c c }
		\toprule
		Modality & Backbone & Model & Top-5 Acc. @ 1s \\
		\midrule
		\multirow{5}{*}{RGB} 
		& RULSTM & TSN & 30.8 \\
		& \bf Ub-RULSTM & TSN &  \bf 31.6 \\
		\cmidrule{2-4}
		& ActionBanks & TSN & 28.6 \\
		& \bf Ub-ActionBanks & TSN & \bf 30.2 \\
		\cmidrule{2-4}
		& DCR & TSM & 33.2 \\
		& \bf Ub-DCR & TSM & \bf 34.2 \\
		\midrule
		\multirow{2}{*}{FLOW} 
		& RULSTM & TSN & 21.4 \\
		& \bf Ub-RULSTM & TSN & \bf 22.9  \\
		\cmidrule{2-4}
		& ActionBanks & TSN & 19.8 \\
		& \bf Ub-ActionBanks & TSN & \bf 20.9 \\
		\midrule
		\multirow{2}{*}{OBJ} 
		& RULSTM & FRCNN & 29.8 \\
		& \bf Ub-RULSTM & FRCNN & \bf 30.7 \\
		\cmidrule{2-4}
		& ActionBanks & FRCNN & 29.1 \\
		& \bf Ub-ActionBanks & FRCNN & \bf 30.9 \\
		\bottomrule
	\end{tabular}
        \vspace{-4mm}
	\label{table_epic55_modality}
\end{table}

\subsubsection{The Chosen Relationships in ConceptNet}
\label{subsec_chosenrelation}

The ConceptNet is a freely-available semantic network, designed to help computers understand the words that people use~\cite{speer2017conceptnet5}.
It contains multiple relationships between words. For better modeling semantic correlations between activity classes, we choose the following meaningful relationships `MotivatedByGoal', `HasPrerequisite', `MannerOf', `UsedFor', `Entails', `LocatedNear', `HasFirstSubevent', `HasSubevent', `HasLastSubevent', `Causes', `CreatedBy', `ReceivesAction', `CausesDesire' and `CapableOf'.

\subsection{Apples-to-Apples Comparison}
\label{subsec_atacomparision}

The results on EK100 and EK55 are shown in~\Cref{table_epic100_modality} and~\Cref{table_epic55_modality}, respectively. We can find that our framework achieves improvement unanimously on different features and different backbones. As shown in~\Cref{table_epic55_modality}, the top-5 accuracy improvement of Ub-RULSM under the FLOW feature is 1.5\%, whereas the top-5 accuracy improvement under the OBJ feature is only 0.9\%. 
This is mainly because the optical flow representation contains more uncertainty brought by noise. 
The effectiveness of our model on different sizes of datasets varies greatly. For example, on the large dataset EK100, Ub-RULSTM has the greatest performance improvement on OBJ feature, but on the small dataset EK55, Ub-RULSTM has the most significant improvement on FLOW feature, because of the weak contribution of the optical flow feature to activity anticipation on large datasets. 
Besides, under the same backbone and features, our proposed framework improves the performance more significantly on EK55 than on EK100. This is mainly due to the large dataset size of the EK100. With the increase of dataset size, the influence of data uncertainty decreases, which leads to the reduction of the performance improvement of our framework.

\begin{table}[t]
	\centering
	\renewcommand{\arraystretch}{0.95}
	\setlength{\tabcolsep}{4mm}
	\caption{The reliability of the uncertainty learning strategy under noisy data. }
    \vspace{-2mm}
		\begin{tabular}{c c c }
			\toprule
			Setting & Top-5 Acc. \% at 1 (s) & DataU \\
			\midrule
			$\eta=0$ & 34.2 & 1.66 \\
			$\eta=1$ & 33.0 & 1.69 \\
			$\eta=5$ & 20.3 & 1.70 \\
			$\eta=10$ & 11.4 & 1.72 \\
			\bottomrule
		\end{tabular}
  \vspace{-2mm}
	\label{table_epic55_ablation_noise_dataset}
\end{table}

\begin{table}[t]
	\centering
	\renewcommand{\arraystretch}{0.95}
	\setlength{\tabcolsep}{3mm}
	\caption{The reliability of the uncertainty learning strategy under high uncertain samples. R=10\% means we remove samples with the highest uncertainty value in the top 10\%, and test on the remaining ones.}
 \vspace{-2mm}
		\begin{tabular}{c c c c c}
			\toprule
			Model & R=0 & R=10\% & R=20\% & R=30\% \\
			\midrule
			RULSTM & 30.8 & 30.7 & 31.8 & 31.7 \\
			Ub-RULSTM & 31.6 & 30.9 & 32.1 & 32.9 \\
			\midrule
			DCR & 33.2 & 35.8 & 38.7 & 44.1 \\
			Ub-DCR & 34.2 & 36.5 & 39.5 & 44.4 \\
			\bottomrule 
		\end{tabular}
  \vspace{-4mm}
	\label{table_epic55_ablation_test_rejection}
\end{table}

\subsection{The Reliability of the Learned Uncertainty Value}

\textbf{The Reliability on Noisy Data.} 
We conduct experiments to assess whether the estimated uncertainty value can capture the `noise' inherent in the data, thus demonstrating the reliability of our framework. We consider samples in EK55 as clean data. Then we deliberately introduce Gaussian noise to pollute them and produce lower-quality samples. 
For the feature tensor $f$, we generate a noise tensor $\epsilon$ and pollute the clean data with $f = f + \eta*\epsilon$, where $\epsilon \sim \mathcal{N}(0, 1)$ and $\eta$ controls the intensity of the pollution. We gradually vary $\eta$ to observe how data uncertainty and anticipation performance evolve. Table~\ref{table_epic55_ablation_noise_dataset} shows that with the increase of $\eta$, the estimated uncertainty of noisy samples increases and the anticipation performance significantly decreases, which collectively indicates the produced uncertainty value can effectively capture the quality of the samples.

\textbf{The Reliability on Samples with High Uncertain.} 
To illustrate the quality of the learned uncertainty value, we investigate how the anticipation performance is affected by the exclusion of samples with high uncertainty values. 
For this analysis, we use the accuracy versus rejection rate metric~\cite{terhorst2020ser,zhang2021relative}. When the accuracy continues to improve as the proportion of removed samples with high uncertainty values increases, this metric indicates that the anticipation model can effectively capture data uncertainty. Hence, we initially sort samples based on their uncertainty values from highest to lowest. Subsequently, we progressively remove a certain proportion of samples with the highest uncertainty values and evaluate the anticipation accuracy using the remaining ones. Table~\ref{table_epic55_ablation_test_rejection} clearly reveals that both Ub-RULSTM and Ub-DCR consistently achieve better performance in all cases, which illustrates that the derived uncertainty value has a stronger correlation with anticipation confidence than other methods. This validates the effect of the uncertainty value in enhancing the overall anticipation performance.

\begin{figure}[t]
	\centering
	\subfigure[The learned uncertainty distribution of Ub-DCR. {\label{vis_uncertainty_dcr}}]
	{\includegraphics[width=0.8\linewidth]{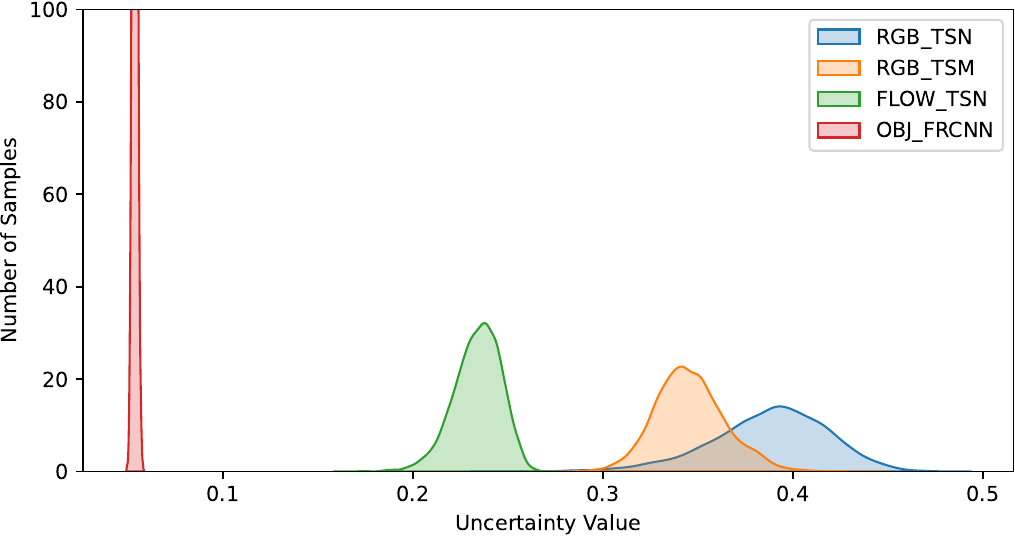}}
	\quad
	\subfigure[The learned uncertainty distribution of Ub-RULSTM. {\label{vis_uncertainty_rulstm}}]
	{\includegraphics[width=0.8\linewidth]{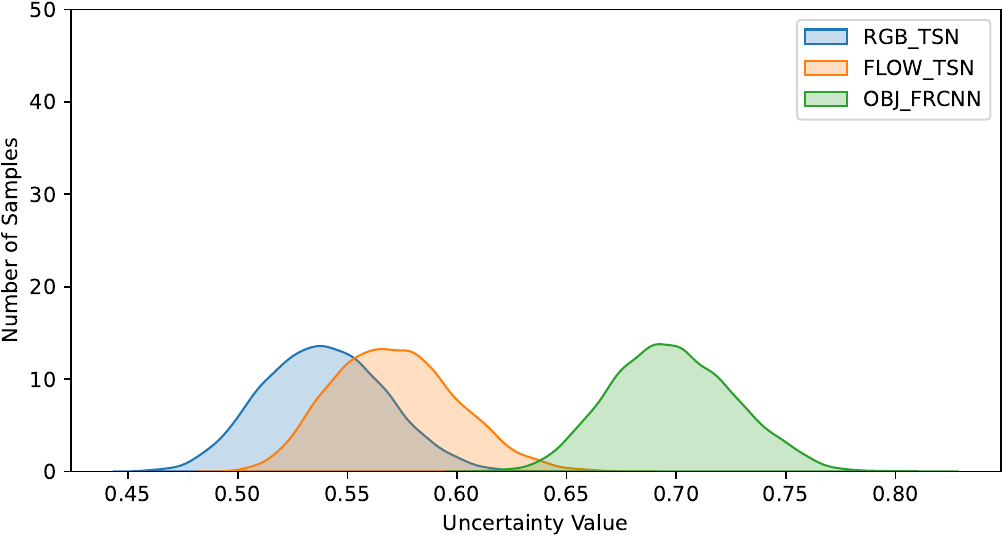}}
	\quad
  \vspace{-4mm}
	\caption{Visualization of the learned uncertainty distribution under different feature modalities on the EK100 validation set.}
 \vspace{-4mm}
	\label{vis_uncertainty}
\end{figure}

\textbf{Understanding the Uncertainty Learning Ability.}
To gain a profound understanding on the uncertainty learning capability, we visualize the learned uncertainty distribution of Ub-DCR and Ub-RULSTM under multiple feature modalities on the EK100 in~\Cref{vis_uncertainty}. For convenient comparison, we normalize the learned uncertainty value to $\left[0, 1\right]$. The vertical axis indicates the number of samples. 
We can see that Ub-DCR generally produces smaller uncertainty values than Ub-RULSTM, due to the strong ability of transformer-based Ub-DCR in modeling long-range feature correlations. Furthermore, we observe that Ub-DCR produces higher uncertainty values for samples under RGB features. Instead, Ub-RULSTM produces higher uncertainty values for samples under OBJ features. 
It is clear that the data uncertainty contained in different feature modalities varies greatly, each backbone exhibits its own distinct strengths in modeling data uncertainty across these diverse feature modalities.

\subsection{Understanding the Robustness of Our Framework}
\label{subsection:robust}

\begin{figure}[t]
	\centering
	\includegraphics[scale=0.4]{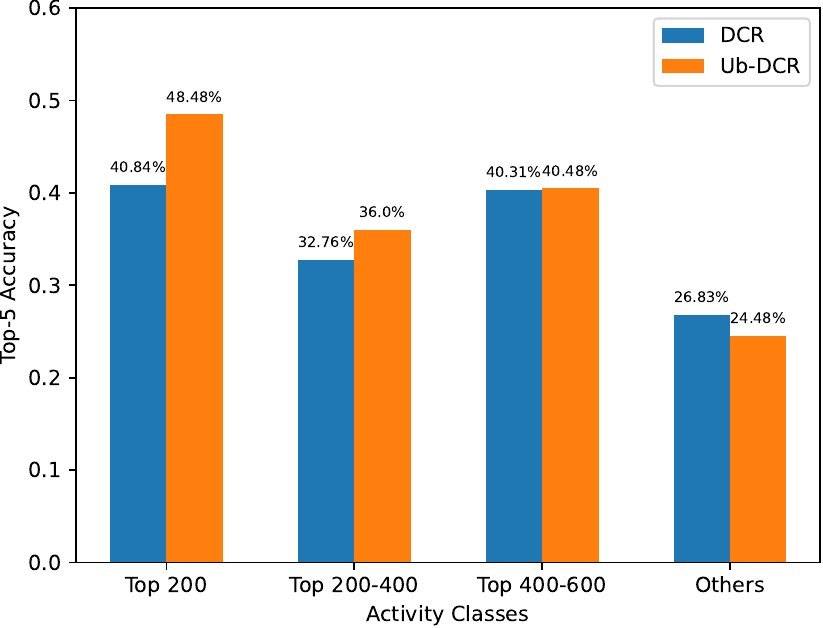}
	\caption{The visualization of the anticipation performance of different methods on uncertain activity classes.}
  \vspace{-4mm}
	\label{fig_robustness_uncertain_classes}
\end{figure}

\textbf{Anticipation Performance on Uncertain Classes.} 
As described in~\Cref{subsec_disadjustment}, we have acquired the internal and external uncertainty matrices of activity classes. For each activity class pair, we first merge their internal and external uncertainty value and then rank them according to their merged values. 
Next, we divide all activity class pairs into four parts and calculate the anticipation performance of DCR and Ub-DCR on each part. The visualization results are shown in~\Cref{fig_robustness_uncertain_classes}. 
On the `Top 200' part, {\it i.e.}, the activity class pairs with the top 200 uncertainty value, Ub-DCR achieves higher anticipation performance than DCR, and the performance gap between DCR and Ub-DCR is also the largest. As the uncertainty value of the activity category decreases, the performance gap between DCR and Ub-DCR decreases. 
This is mainly because the possibility of co-occurrence between activity categories gradually decreases. 
If there is a strong co-occurrence between activity categories or large activity evolution uncertainty, that is, given videos containing the same activity category, they may evolve into any activity in the category pair, the Ub-DCR can accurately correlate labels with valid features through~\Cref{eqa_target_label} and avoid the confusion caused by one-hot labels. 
These observations show that our framework can effectively deal with the activity categories with high uncertainty values.

\begin{figure}[t]
	\centering
	\includegraphics[scale=0.4]{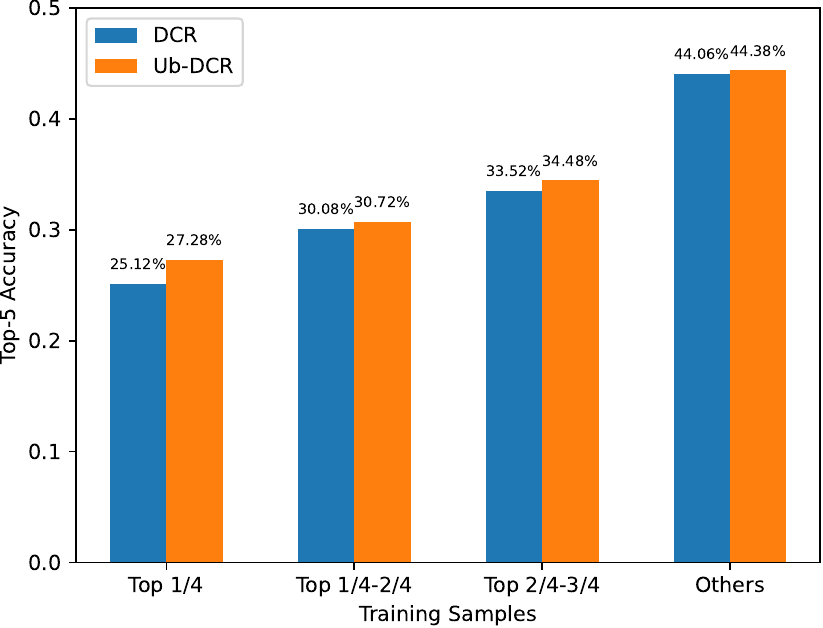}
	\caption{The visualization of the anticipation performance of different methods with respect to the sample uncertainty.}
  \vspace{-4mm}
	\label{fig_robustness_uncertain_samples}
\end{figure}

\textbf{Anticipation Performance on Uncertain Samples.}
We first rank samples based on their uncertainty value produced by Ub-DCR. 
Then, we divide all samples into four parts and visualize the anticipation performance of DCR and Ub-DCR on each part in~\Cref{fig_robustness_uncertain_samples}. 
On the `Top 1/4' part, {\it i.e.}, the samples with the uncertainty value in the top 1/4, Ub-DCR achieves higher anticipation performance than DCR, and the performance gap between DCR and Ub-DCR is the largest. 
The performance difference between DCR and Ub-DCR decreases with the decrease of the sample uncertainty value. 
These observations show that our framework has good robustness and can effectively deal with samples with high uncertainty values.

\begin{table}[t]
	\centering
	\renewcommand{\arraystretch}{0.95}
	\setlength{\tabcolsep}{1.6mm}
	\caption{Activity anticipation comparison on the Tail Classes Subset from the EK100 validation set.}
  \vspace{-2mm}
	\begin{tabular}{c c c c c c c}
		\toprule
		\multirow{2}{*}{Model} & \multicolumn{3}{c}{Overall \% @ 1s} & \multicolumn{3}{c}{Tail \% @ 1s} \\
		\cmidrule{2-7}
		& Verb & Noun & Act & Verb & Noun & Act \\
		\midrule
		DCR~\cite{xu2022learning} & 31.0 & 31.1 & 14.6 & 26.1 & 25.8 & 12.5 \\
		DCR+CB & \bf 31.7 & 31.4 & 12.7 & 26.6 & 26.4 & 11.4 \\
		\bf Ub-DCR & 31.4 & \bf 31.8 & \bf 14.8 & \bf 26.8 & \bf 27.5 & \bf 13.5 \\
		\bottomrule
	\end{tabular}
  \vspace{-2mm}
	\label{table_robustness_longtail}
\end{table}

\begin{figure}[t]
	\centering
	\includegraphics[scale=0.27]{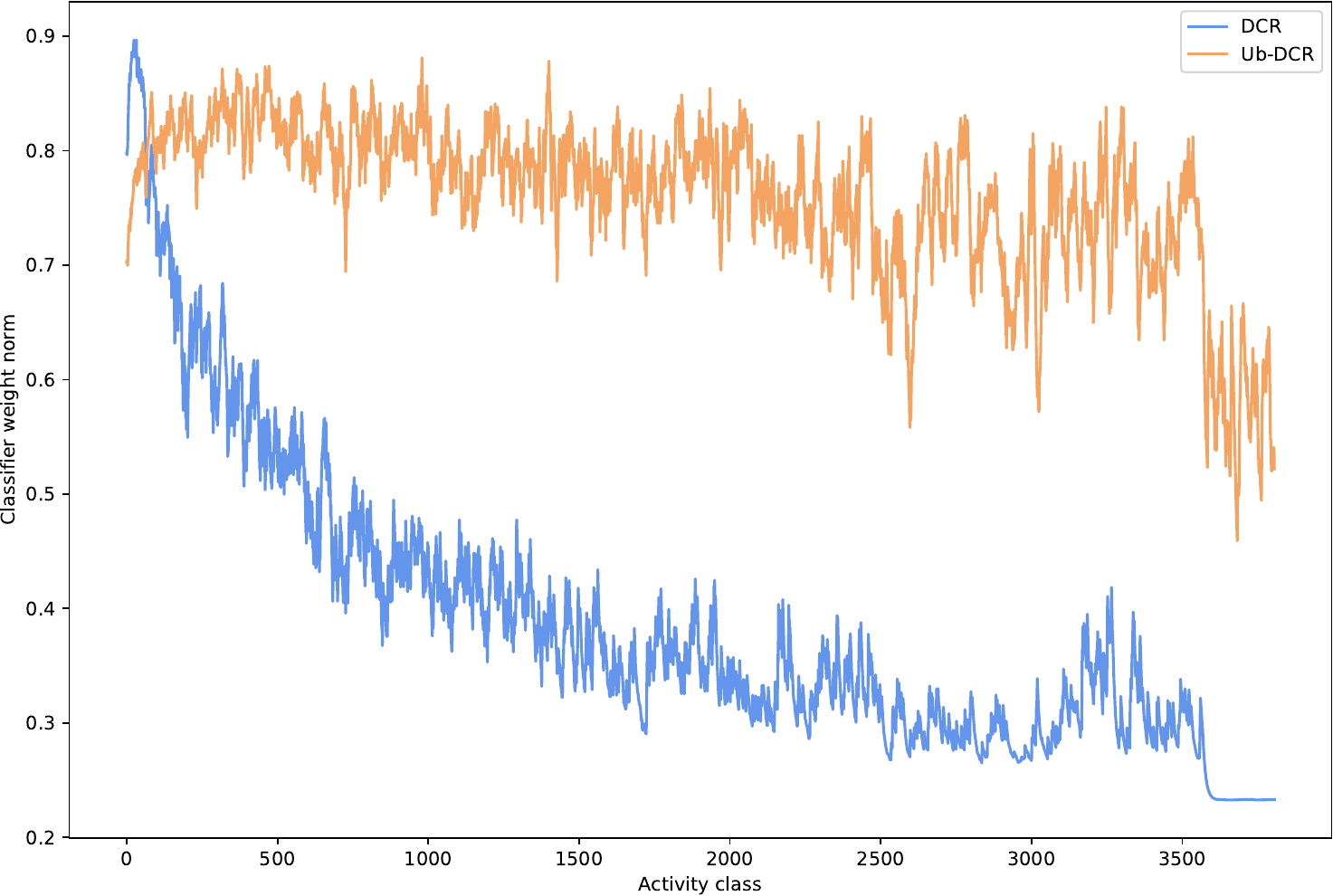}
	\caption{The weight-norms of target activity classifier of different backbones on the EK100. The activity classes are sorted by descending order w.r.t. the number of samples.}
  \vspace{-4mm}
	\label{fig_robustness_tailclass_weight}
\end{figure}

\textbf{Anticipation Performance on Tail Classes.} 
As shown in~\Cref{table_epic100_validation} and~\Cref{table_epic100_test}, the relative anticipation performance improvement of our method on the Tail Classes subset of the EK100 is much higher than that on the whole validation (or test) set or the Unseen Participants subset. 
For example, Ub-DCR improves the anticipation performance by 47.6\% over DCR$^\dag$ on the Tail Classes subset 'Act', while Ub-DCR only improves the performance by 26.9\% over DCR$^\dag$ on 'Act' of the whole test set. 
This indicates that our method is capable of dealing with data with long-tail distribution issues. 
To comprehensively verify this, we visualize the target activity classifier weight-norms of DCR and Ub-DCR for the EK100. 
We sort the target activity categories in descending order with respect to their number of instances and show the L$_2$-norms of their weight vectors in~\Cref{fig_robustness_tailclass_weight}. 
Compared with DCR, the weight-norms of the tail classes tend to be larger and the weight-norms of the head classes are relatively smaller for Ub-DCR. 
This means our framework reaches a better trade-off between head and tail classes.

\begin{table*}[t]
	\centering
	\renewcommand{\arraystretch}{0.95}
	\setlength{\tabcolsep}{1.6mm}
	\caption{Activity anticipation results on the EK100 validation set.}
        \vspace{-2mm}
		\begin{tabular}{c c c c c c c c c c c}
			\toprule
			\multirow{2}{*}{Model} & \multirow{2}{*}{Modality} & \multicolumn{3}{c}{Overall \% @ 1s} & \multicolumn{3}{c}{Unseen \% @ 1s} & \multicolumn{3}{c}{Tail \% @ 1s} \\
			\cmidrule{3-11}
			&  & Verb & Noun & Act & Verb & Noun & Act & Verb & Noun & Act \\
			\midrule
			ActionBanks~\cite{sener2020temporal} & RGB, OBJ, FLOW, ROI & 23.2 & 31.4 & 14.7 & 28.0 & 26.2 & 14.5 & 14.5 & 22.5 & 11.8 \\
			AVT-TSN~\cite{activipative_transformer} & RGB, OBJ & 25.5 & 31.8 & 14.8 & 25.5 & 23.6 & 11.5 & 18.5 & 25.8 & 12.6 \\
			AVT~\cite{activipative_transformer} & RGB, OBJ & 28.2 & 32.0 & 15.9 & 29.5 & 23.9 & 11.9 & 21.1 & 25.8 & 14.1 \\
			MeMViT,32$\times$3~\cite{wu2022memvit} & RGB & 32.2 & 37.0 & 17.7 & 28.6 & 27.4 & 15.2 & 25.3 & 31.0 & 15.5 \\
			\midrule
			RULSTM~\cite{rulstm} & RGB, OBJ, FLOW & 27.8 & 30.8 & 14.0 & 28.8 & 27.2 & 14.2 & 19.8 & 22.0 & 11.1 \\
			\bf Ub-RULSTM & RGB, OBJ, FLOW & \bf 30.4 & \bf 34.8 & \bf 16.6 & \bf 33.8 & \bf 27.5 & \bf 14.5 & \bf 23.4 & \bf 28.8 & \bf 14.5 \\
			\midrule
			DCR~\cite{xu2022learning} & RGB, OBJ & 32.6 & 36.9 & 18.3 & 31.8 & 25.8 & 14.7 & 26.6 & 30.1 & 15.8 \\
			\bf Ub-DCR & RGB, OBJ & \bf 35.1 & \bf 40.0 & \bf 19.1 & \bf 36.2 & \bf 26.7 & \bf 15.9 & \bf 29.6 & \bf 34.8 & \bf 17.8 \\
			\bottomrule
		\end{tabular}
	\label{table_epic100_validation}
	\vspace{-2mm}
\end{table*}

\begin{table}[t]
	\centering
	\renewcommand{\arraystretch}{0.95}
	\setlength{\tabcolsep}{2mm}
	\caption{Study on the effect of model uncertainty of our framework by controlling the training set size.}
 \vspace{-2mm}
		\begin{tabular}{c c c }
			\toprule
			Training Set Size & Top-5 Acc. @ 1s & DataU \\
			\midrule
			EPIC55 1/4 & 29.1 & 1.61 \\
			EPIC55 2/4 & 31.9 & 1.60 \\
			EPIC55 3/4 & 32.8 & 1.65 \\
			EPIC55 4/4 & 34.2 & 1.66 \\
			\bottomrule
		\end{tabular}
  \vspace{-4mm}
	\label{table_epic55_ablation_training_set_size}
\end{table}

Furthermore, we apply the class-balanced loss CB~\cite{cui2019class} on DCR to compare the ability of different methods on long-tail distribution problems. The results are shown in~\Cref{table_robustness_longtail}. 
We can find that Ub-DCR achieves better performance than DCR+CB on the Tail Classes subset. Compared with one-hot labels used by existing anticipation methods, our strategy ensures that the constructed label contains both the original target activity and other potential co-occurred activities. 
This somewhat relates minority activity classes to majority classes, and solves the long-tail distribution problem from a data sampling perspective. 
Besides, we can see that the performance of DCR+CB is worse than DCR. 
This is mainly due to the construction of the activity category. In EK100, activity categories are composed of all unique (verb, noun) class pairs, resulting in various categories with a sample size of 1 or 0. However, the loss weight of the class-balanced loss CB is adjusted according to the number of category samples to deal with long-tail distribution issues. Under this situation, many classical algorithms cannot address the long-tail problem in activity anticipation task datasets.

\subsection{Jointly Exploring the Data and Model Uncertainty}

While our framework is specially designed to model data uncertainty, it is important to acknowledge that the outputs may contain model uncertainty. We will explore the proportion of the model uncertainty within predictive uncertainty and the impact of model uncertainty on the final results.

\textbf{Quantifying Model Uncertainty with Dataset Control.} 
Since model uncertainty can be largely mitigated with sufficient data~\cite{hora1996aleatory, der2009aleatory, hullermeier2021aleatoric}, we begin with the insight that model uncertainty varies with training data size while data uncertainty does not~\cite{kendall2017uncertainties}. Specifically, we assess the proportion of model uncertainty and analyze the changes of predicted uncertainty values under different training set sizes. 
We vary the training set size of EK55 to 3/4, 1/2, and 1/4, then summarize the variations of uncertainty value and anticipation performance of Ub-DCR in Table~\ref{table_epic55_ablation_training_set_size}. 
We can see the uncertainty value increases slightly as the training set size increases, which is negligible compared to the absolute data uncertainty value. Obviously, the predicted uncertainty contains only a marginal amount of model uncertainty.

\textbf{Quantifying Model Uncertainty Using Monte Carlo Dropout.} We employ the widely-used Monte Carlo Dropout (MCDO) method~\cite{gal2016dropout, kendall2017uncertainties, depeweg2018decomposition} to explicitly study the correlation between data and model uncertainty. MCDO is a simple but efficient method to capture model uncertainty. By activating the dropout layers at test time and performing $T$ forward passes on a given sample set, one can obtain multiple sets of probability distributions regarding the anticipation results. This allows us to estimate model uncertainty using operations such as mean and entropy. 
We apply the MCDO approach to Ub-DCR on EKI55 and EK100, and summarize the changes of data and model uncertainty as well as their impact on the anticipation performance in Table~\ref{table_epic55_modality_mcdropout}. We use 50 samples for Monte Carlo Dropout. We can see that data uncertainty remains relatively stable, while model uncertainty does exist. Furthermore, the simultaneous examination of model and data uncertainty can lead to further improvements on anticipation performance in EK55. However, this is not the case in EK100, which might be attributed to a diminishing influence of model uncertainty as the dataset size increases. Since the model uncertainty can be painlessly modeled using MCDO and can be removed easily in application scenarios~\cite{kendall2017uncertainties, ayhan2022test, hullermeier2021aleatoric, gruber2023sources} involving big data and deep neural networks, this paper focuses on the challenge of data uncertainty learning in priority.

\begin{table}[t]
	\centering
	\renewcommand{\arraystretch}{0.95}
	\setlength{\tabcolsep}{1mm}
	\caption{Apples-to-apples comparison of data and model uncertainty on the EK55 and EK100 validation set. `-MCDO' means injecting Monte Carlo Dropout into our framework under different backbones.}
 \vspace{-2mm}
	\begin{tabular}{c c c c c c}
		\toprule
		Dataset & Backbone & Model & Top-5 & DataU & ModelU \\
		\midrule
		\multirow{2}{*}{EK55} 
		& Ub-DCR & TSM & 34.2 & 1.66 & / \\
		& Ub-DCR-MCDO & TSM & 34.7 & 1.69 & 0.25 \\
		\midrule
		\multirow{4}{*}{EK100} 
		& Ub-DCR & TSN & 14.8 & 0.39 & / \\
		& Ub-DCR-MCDO & TSN & 14.7 & 0.40 & 0.27 \\
		\cmidrule{2-6}
		& Ub-DCR & TSM & 16.2 & 0.35 & / \\
		& Ub-DCR-MCDO & TSM & 16.1 & 0.35 & 0.27 \\
		\bottomrule
	\end{tabular}
 \vspace{-4mm}
	\label{table_epic55_modality_mcdropout}
\end{table}


\begin{table*}[t]
	\centering
	\renewcommand{\arraystretch}{0.95}
	\setlength{\tabcolsep}{1.6mm}
	\caption{Activity anticipation results on the EK100 test set.}
 \vspace{-2mm}
		\begin{tabular}{c c c c c c c c c c c}
			\toprule
			\multirow{2}{*}{Model} & \multirow{2}{*}{Modality} & \multicolumn{3}{c}{Overall \% @ 1s} & \multicolumn{3}{c}{Unseen \% @ 1s} & \multicolumn{3}{c}{Tail \% @ 1s} \\
			\cmidrule{3-11}
			&  & Verb & Noun & Act & Verb & Noun & Act & Verb & Noun & Act \\
			\midrule
			RULSTM~\cite{rulstm} & RGB & 24.7 & 26.4 & 10.5 & 17.9 & 23.2 & 9.1 & 17.3 & 16.8 & 7.4 \\
			MM-TBN~\cite{zatsarynna2021multi} & RGB, OBJ, FLOW & 21.5 & 26.8 & 11.0 & 20.8 & 28.3 & 12.2 & 13.2 & 15.4 & 7.2 \\
			ActionBanks~\cite{sener2020temporal}  & RGB, OBJ, FLOW, ROI & / & / & 12.6 & / & / & 10.5 & / & /& 8.9 \\
			AVT~\cite{activipative_transformer} & RGB, OBJ & 25.6 & 28.8 & 12.6 & 20.9 & 22.3 & 8.8 & 19.0 & 22.0 & 10.1 \\
			DCR~\cite{xu2022learning}  & RGB, OBJ & / & / & 17.3 & / & / & 14.1 & / & /& 14.3 \\
			\midrule
			RULSTM~\cite{rulstm} & RGB, OBJ, FLOW& 25.3 & 26.7 & 11.2 & 19.4 & 26.9 & 9.7 & 17.6 & 16.0 & 7.9 \\
			\bf Ub-RULSTM & RGB, OBJ, FLOW & \bf 27.3 & \bf 30.2 & \bf 13.0 & \bf 20.3 & \bf 27.3 & \bf 10.6 & \bf 20.2 & \bf 21.7 & \bf 10.6 \\
			\midrule
			DCR$^\dag$~\cite{xu2022learning} & RGB, OBJ & 27.0 & 31.4 & 11.9 &22.8 &  30.7 & 11.0 & 21.0 & 22.5 & 8.4 \\
			\bf Ub-DCR & RGB, OBJ & \bf 29.4 & \bf 32.9 & \bf 15.1 &  \bf 23.5 & \bf  31.2 & \bf 11.5 & \bf 23.5 & \bf 25.3 & \bf 12.4 \\
			\bottomrule
		\end{tabular}
  \vspace{-2mm}
	\label{table_epic100_test}
\end{table*}

\begin{table}[t]
	\centering
	\renewcommand{\arraystretch}{0.95}
	\setlength{\tabcolsep}{2mm}
	\caption{Activity anticipation results on the EGTEA Gaze+.}
 \vspace{-2mm}
		\begin{tabular}{c c c c c c c}
			\toprule
			\multirow{2}{*}{ Model} & \multicolumn{6}{c}{Top-5 Accuracy \% at different $\tau_a$ (s)} \\
			\cmidrule{2-7}
			& 1.5 & 1.25 & 1.0 & 0.75 & 0.5 & 0.25 \\
			\midrule
			DMR~\cite{vondrick2016anticipating}& / & / & 55.7 & / & / & / \\
			ATSN~\cite{damen2018scaling}& / & / & 40.5 & / & / & / \\
			MCE~\cite{furnari2018leveraging}& / & / & 56.3 & / & / & / \\
			ED~\cite{gao2017red} & 46.9 & 48.4 & 50.2 & 51.9 & 50.0 & 49.2 \\
			FN~\cite{de2018modeling} & 56.8 & 58.3 & 60.1 & 62.0 & 64.0 & 66.5 \\
			RL~\cite{ma2016learning} & 58.2 & 60.4 & 62.6 & 64.7 & 67.4 & 70.4 \\
			EL~\cite{jain2016recurrent} & 59.8 & 61.6 & 64.6 & 66.9 & 69.6 & 72.4  \\
			ImagineRNN~\cite{wu2020learning}& / & / & 66.7 & 68.5 & 72.3 & 74.6 \\
			SRL~\cite{srl} & 64.9 & 66.5 & 70.7 & 73.5 & 78.0 & 82.6 \\
			MGRKD~\cite{huang20multimodal} & 65.2 & 67.7 & 70.9 & 74.3 & 77.5 & 79.6 \\
			\midrule
			RULSTM~\cite{rulstm} & 61.4 & 63.5 & 66.4 & 68.4 & 71.8 & 74.3 \\
			\bf Ub-RULSTM & \bf 61.7 & \bf 64.1 & \bf 67.9 & \bf 69.5 & \bf 72.4 & \bf 74.6 \\
			\midrule
			DCR~\cite{xu2022learning} & / & / & 67.9 &  / & / & /  \\
			\bf Ub-DCR & / & / & \bf 68.2 & / & / & / \\
			\bottomrule
		\end{tabular}
  \vspace{-4mm}
	\label{table_gteagazeplus}
\end{table}

\subsection{Comparison with the State-of-the-Art}
\label{subsec_comparsion}

\textbf{EPIC-KITCHENS-100.} The comparison results on the validation set and test set are shown in~\Cref{table_epic100_validation} and~\Cref{table_epic100_test}. 
As shown in~\cite{xu2022learning}, DCR fuses the results of models under RGB feature from TSM and TSN, and OBJ feature from FRCNN with weight 1:1:1 on the validation set. The results of DCR on the test set are obtained by fusing the model output under RGB feature from TSM and TSN, and the AVT model with weight 1:0.5:1. Since the AVT model combines the predictions of multiple AVT model variants and multiple backbones, they are difficult to reproduce. 
Hence, we capture the anticipation results by late-fusion of the output of Ub-DCR under RGB feature from TSM and TSN, and OBJ feature from FRCNN with weight 1:1:1 on the validation and test sets. For a fair comparison, we also implement DCR$^\dag$ on the test set with the same setting as Ub-DCR. 

As shown in~\Cref{table_epic100_validation} and~\Cref{table_epic100_test}, our framework significantly improves the anticipation performance on RULSTM and DCR backbones in the model ensemble setting. 
By comparing the anticipation results in~\Cref{table_epic100_modality}, \Cref{table_epic100_validation} and~\Cref{table_epic100_test}, the improvement achieved by our framework is even more prominent under the ensemble setting. As analyzed above, the data uncertainty contained in each feature modality is different. Hence, modeling data uncertainty from multiple feature modalities ensures better anticipation performance. 
Besides, we notice that the performance of RULSTM is not as good as ActionBanks, because ActionBanks uses additional ROI features. On the contrary, Ub-RULSTM achieves better performance than ActionBanks.  
The Ub-DCR achieves higher anticipation performance than other competitors on the validation set. It is worth noting that both AVT and MeMViT are models trained on raw video data, while we only use pre-extracted features. Besides, the performance improvement on the Unseen participants subset demonstrates the generalizability of our framework even under the `zero-shot' settings.

\begin{table}[t]
	\centering
	\renewcommand{\arraystretch}{0.95}
	\setlength{\tabcolsep}{1.2mm}
	\caption{Activity anticipation results on the MECCANO test set.}
  \vspace{-2mm}
	\begin{tabular}{c c c c c c c c c}
		\toprule
		\multirow{2}{*}{Model} & \multicolumn{8}{c}{Top-5 Accuracy \% at different $\tau_a$ (s)} \\
		\cmidrule{2-9}
		& 2 & 1.75 & 1.5 & 1.25 & 1.0 & 0.75 & 0.5 & 0.25 \\
		\midrule
		RULSTM~\cite{rulstm} & 54.7 & 56.0 & 56.6 & 57.7 & 58.2 & 60.0 & 61.3 & 63.4 \\
		\bf Ub-RULSTM & \bf 60.3 & \bf 61.5 & \bf 61.2 & \bf 62.3 & \bf 62.7 & \bf 63.9 & \bf 64.0 & \bf 65.7 \\
		\midrule
		DCR~\cite{xu2022learning} & / & / & / & / & 56.7 & / & / & / \\
		\bf Ub-DCR & / & / & / & / & \bf 60.3 & / & / & / \\
		\bottomrule
	\end{tabular}
  \vspace{-2mm}
	\label{table_meccano}
\end{table}

\begin{table}[t]
	\renewcommand{\arraystretch}{1}
	\setlength{\tabcolsep}{0.35mm}
	\centering
	\caption{Third-person activity anticipation results on 50 Salads.}
  \vspace{-2mm}
	\begin{tabular}{c c c c c c c c c}
		\toprule
		Dataset & \multicolumn{8}{c}{50 Salads} \\
		\midrule
		Observed & \multicolumn{4}{c}{20 \%} & \multicolumn{4}{c}{30 \%} \\
		\midrule
		Predicted & 10 \% & 20 \% & 30 \% & 50 \% & 10 \% & 20 \% & 30 \% & 50 \% \\
		\midrule
		RNN model~\cite{abu2018will} & 30.1 & 25.4 & 18.7 & 13.5 & 30.8 & 17.2 & 14.8 & 09.8 \\
		Time-cond.~\cite{ke2019time} & 32.5 & 27.6 & 21.3 & 16.0 & 35.1 & 27.1 & 22.1 & 15.6 \\
		ActionBanks~\cite{sener2020temporal} & 25.5 & 19.9 & 18.2 & 15.1 & 30.6 & 22.5 & 19.1 & 11.2 \\					
		CycleCons~\cite{abu2021long} & 34.8 & 28.4 & 21.8 & 15.3 & 34.4 & 23.7 & 19.0 & 15.9 \\
		A-ACT~\cite{Gupta2022action} & 35.4 & 29.6 & 22.5 & 16.1 & 35.7 & 25.3 & 20.1 & 16.3 \\
		SRL~\cite{srl} & 37.9 & 28.8 & 21.3 & 11.1 & 37.5 & 24.1 & 17.1 & 09.1 \\
		\midrule
		FUTR~\cite{gong2022future} & 39.6 & 27.5 & 23.3 & 17.8 & 35.2 & 24.9 & 24.2 & 15.3 \\
		\bf Ub-FUTR & \bf 40.7 & \bf 28.5 & \bf 24.2 & \bf 18.4 & \bf 36.3 & \bf 25.5 & \bf 24.9 & \bf 16.2 \\
		\bottomrule
	\end{tabular}
  \vspace{-4mm}
	\label{table_50salads_results}
\end{table}

\textbf{EGTEA Gaze+.} 
As shown in~\Cref{table_gteagazeplus}, the activity anticipation performance of Ub-RULSTM and Ub-DCR are both higher than RULSTM and DCR at all anticipation times.

\textbf{MECCANO.} 
As shown in~\Cref{table_meccano}, the anticipation performance of Ub-RULSTM and Ub-DCR are both higher than RULSTM and DCR at all anticipation time settings. This indicates that our framework is also effective on videos exhibiting human behaviors in industry scenarios beyond kitchen activities.

\textbf{50 Salads.} We conduct experiments under a long-term anticipation setting. Following~\cite{abu2018will}, the input is a particular percentage ({\it i.e.}, 20\%) of each video, and the goal is to anticipate the activities of the following sub-sequence with a percentage ({\it i.e.}, 10\%, 20\%, 30\% and 50\%) of the video. As shown in Table~\ref{table_50salads_results}, the performance of Ub-FUTR is higher than FUTR at most anticipation timestamps. This indicates that our framework is also effective in third-person video activities under a long-term anticipation setting.

\begin{table*}[t]
	\centering
	\renewcommand{\arraystretch}{0.95}
	\caption{Activity anticipation results on the EK55 validation set.}
  \vspace{-2mm}
	\resizebox{\linewidth}{!}{
		\begin{tabular}{c c c c c c c c c c c c c c c}
			\toprule
			\multirow{2}{*}{Model} & \multicolumn{8}{c}{Top-5 Accuracy \% at different $\tau_a$ (s)} & \multicolumn{3}{c}{Top-5 Acc. \% @ 1s} & \multicolumn{3}{c}{M Top-5 Rec. \% @ 1s}\\
			\cmidrule{2-15}
			& 2 & 1.75 & 1.5 & 1.25 & 1.0 & 0.75 & 0.5 & 0.25 & Verb & Noun & Act & Verb & Noun & Act \\
			\midrule 
			MCE~\cite{furnari2018leveraging} & / & / & / & / & 26.1 & / & / & / & 73.4 & 38.9 & 26.1 & 34.6 & 32.6 & 6.5 \\
			VN-CE~\cite{damen2018scaling} & / & / & / & / & 17.3 & / & / & / & 77.7 & 39.5 & 17.3 & 34.1 & 34.5 & 7.7 \\
			FHOI(I3D) \cite{liu2020forecasting} & / & / & / & / & 25.5 & / & / & / & 76.5 & 42.6 & 25.5 & / & / & / \\
			VNMCE+T5~\cite{furnari2018leveraging} & / & / & / & / & 26.0 & / & / & / & 74.1 & 39.1 & 26.0 & 41.6 & 35.5 & 5.8 \\
			ED~\cite{gao2017red} & 21.5 & 22.2 & 23.2 & 24.8 & 25.8 & 26.7 & 27.7 & 29.7 & 75.5 & 43.0 & 25.8 & 41.8 & 42.6 & 11.0 \\
			FN~\cite{de2018modeling} & 23.5 & 24.1 & 24.7 & 25.7 & 26.3 & 26.9 & 27.9 & 29.0 & 74.8 & 40.9 & 26.3 & 35.3 & 37.8 & 6.6 \\
			RL~\cite{ma2016learning} & 26.0 & 26.5 & 27.2 & 28.5 & 29.6 & 30.8 & 31.9 & 32.8 & 76.8 & 44.5 & 29.6 & 40.8 & 40.9 & 10.6 \\
			EL~\cite{jain2016recurrent} & 24.7 & 25.7 & 26.4 & 27.4 & 28.6 & 30.3 & 31.5 & 33.6 & 75.7 & 43.7 & 28.6 & 38.7 & 40.3 & 8.6 \\
			HORST~\cite{tai2021higher} & 25.5 & 26.4 & 27.8 & 29.2 & 30.7 & 31.5 & 32.5 & 33.5 & 77.7 & 46.3 & 30.7 & 36.5 & 44.3 & 10.9 \\
			SRL~\cite{srl} & 30.2 & 31.3 & 32.4 & 34.1 & 35.5 & 36.8 & 38.6 & 40.5  & / & / & 35.5 & / & / & / \\
			ImagineRNN~\cite{wu2020learning} & / & / & 32.5 & 33.6 & 35.6 & 36.7 & 38.5 & 39.4  & / & / & 35.6 & / & / & / \\
			ActionBanks~\cite{sener2020temporal} & 30.9 & 31.8 & 33.7 & 35.1 & 36.4 & 37.2 & 39.5 & 41.3 & / &  / & 35.6 & / & /  & / \\
			AVT+~\cite{activipative_transformer} & / & / & / & / & 37.6 & / & / & /& / & / & 37.6 & / & / & / \\
			DCR~\cite{xu2022learning} & / & / & / & / & 41.2 & / & / & / & 81.8 & 60.8 &  41.2 & / & / & / \\
			\midrule
			RULSTM~\cite{rulstm} & 29.5 & 30.8 & 32.2 & 33.4 & 35.3 & 36.3 & 37.4 & 39.0 & 79.6 & 51.8 & 35.3  & 43.8 & 49.9 & 15.1 \\
			\bf Ub-RULSTM  & \bf 30.1 & \bf 31.5 & \bf 33.1 & \bf 34.3 & \bf 35.8 & \bf 36.8 & \bf 38.4 & \bf 39.9 & \bf 80.4 & \bf 53.5 & \bf 35.8 & \bf 44.8 & \bf 53.0 & \bf 16.0 \\
			\midrule
			DCR(TSM)~\cite{xu2022learning} & / & / & / & / & 33.2 & / & / & / & 67.4 & 53.3 &  33.2 & / & / & / \\
			\bf Ub-DCR(TSM)  & / & / & / & / & \bf 34.2 & / & / & / & \bf 78.2 & \bf 55.2 & \bf 34.2 & / & / & / \\
			\bottomrule
	\end{tabular}}
  \vspace{-2mm}
	\label{table_epic55_validation}
\end{table*}

\textbf{EPIC-KITCHENS-55.} 
The comparison results on the validation and test sets are shown in~\Cref{table_epic55_validation} and~\Cref{table_epic55_test}. 
In~\Cref{table_epic55_validation}, the anticipation performance of Ub-RULSTM (or Ub-DCR) is significantly improved at most anticipation timestamps. 
Our framework also improves the performance of each backbone on the test set. Since the test set \textbf{S2} contains kitchens that are not present in the training set, this improvement shows that our method can generalize to unseen scenes in video event anticipation.

Furthermore, by using the specially designed weights to fuse the results of different models, DCR achieves higher performance than other methods on validation and test sets. Concretely, as shown in~\cite{xu2022learning}, on the validation set, DCR fuses the anticipation results of models trained on the RGB feature from TSM, TSN and irCSN152, and OBJ feature from FRCNN with weight 1:1:1:1. On the test set, the results of DCR are obtained by fusing the model predictions under the RGB feature from TSM and irCSN152, and the ensemble results of AVT model with weight 1:1:1 for test set \textbf{S1} and 0.5:1.5:1.5 for test set \textbf{S2}. 
As the effectiveness of Ub-DCR has been proved in the model ensemble setting in~\Cref{table_epic100_validation} and~\Cref{table_epic100_test}, we only compare Ub-DCR with DCR under RGB feature from TSM model. The experimental results show that our framework is still effective.

\begin{table}[t]
	\centering
	\renewcommand{\arraystretch}{0.95}
	\setlength{\tabcolsep}{1.3mm}
	\caption{Ablation studies on each component of the framework.}
  \vspace{-2mm}
	\begin{tabular}{c c c c c c c c c}
		\toprule
		\multirow{2}{*}{Setting} & \multicolumn{8}{c}{Top-5 Accuracy \% at different $\tau_a$ (s)} \\
		\cmidrule{2-9}
		& 2 & 1.75 & 1.5 & 1.25 & 1.0 & 0.75 & 0.5 & 0.25 \\
		\midrule
		\bf Ub-Baseline & \bf 27.1 & \bf 28.5 & \bf 29.2 & \bf 30.5 & \bf 32.1 & \bf 33.7 & \bf 34.7 & \bf 36.1 \\
		\midrule
		-disadj & 26.9 & 27.7 & 28.8 & 29.9 & 31.6 & 32.2 & 33.6 & 35.2 \\
		-optlabel & 26.2 & 27.0 & 28.5 & 29.7 & 31.2 & 32.5 & 33.8 & 35.4 \\
		-srul & 26.7 & 27.5 & 28.8 & 30.0 & 32.0 & 32.4 & 34.2 & 35.5 \\
		-trul & 26.3 & 27.4 & 28.7 & 29.8 & 31.6 & 33.1 & 34.1 & 35.8 \\
		\midrule
		Baseline & 26.2 & 26.8 & 28.2 & 29.7 & 30.9 & 32.1 & 33.3 & 34.0 \\
		\bottomrule
	\end{tabular}
  \vspace{-4mm}
	\label{table_epic55_ablation_rgb}
\end{table}

\subsection{Ablation Studies}
\label{subsec_ablation}

The ablation studies are conducted on RGB feature from TSN model on the EK55 validation set. The results are shown in~\Cref{table_epic55_ablation_rgb}, where `disadj' and `optlabel' mean the distribution adjustment strategy and the construction of more precise target activity labels, respectively. `srul' and `trul' indicate the sample-wise and temporal relative uncertainty learning strategies, respectively.

\textbf{The Effect of the Distribution Adjustment.} 
By comparing the results of the Ub-Baseline and the row `-disadj' in~\Cref{table_epic55_ablation_rgb}, the performance decreases significantly without distribution adjustment, which indicates the effectiveness of using uncertainty value to adjust the generated probability distribution of the target activity categories.

\begin{table}[t]
	\centering
	\renewcommand{\arraystretch}{0.95}
	\setlength{\tabcolsep}{0.5mm}
	\caption{Activity anticipation results on the EK55 test set.}
 \vspace{-2mm}
		\begin{tabular}{c c c c c c c c}
			\toprule
			\multirow{2}{*}{Setting} & \multirow{2}{*}{Model} & \multicolumn{3}{c}{Top-1 Acc. @ 1s} & \multicolumn{3}{c}{Top-5 Acc. @ 1s} \\
			\cmidrule{3-8}
			& & Verb & Noun & Act &  Verb & Noun & Act \\
			\midrule
			\multirow{5}{*}{\textbf{S1}}
			& ATSN~\cite{damen2018scaling} & 31.8 & 16.2 & 6.0 & 76.6 & 42.2 & 28.2 \\
			& MCE~\cite{furnari2018leveraging} & 27.9 & 16.1 & 10.8 & 73.6 & 39.3 & 25.3 \\
			& ActionBanks~\cite{sener2020temporal} & 31.4 & 22.6 & 16.4 & 75.2 & 47.2 & 36.4 \\
			& SRL~\cite{srl} & 34.9 & 22.8 & 14.2 & 79.6 & 52.0 & 34.6 \\
			& ImagineRNN~\cite{wu2020learning} & 35.4 & 22.8 & 14.7 & 79.7 & 52.1 & 35.0 \\
			& MMTCN-TBN~\cite{zatsarynna2021multi} & 37.2 & 23.7 & 15.4 & 79.5 & 51.9 & 34.4 \\
			& FHOI~\cite{liu2020forecasting} & 35.0 & 20.9 & 14.0 & 77.1 & 46.5 & 31.3 \\
			& Ego-OMG~\cite{dessalene2021forecasting} & 32.2 & 24.9 & 16.0 & 77.4 & 50.2 & 34.5 \\
			& MGRKD~\cite{huang20multimodal} & 38.7 & 25.2 & 17.0 & 79.2 & 53.4 & 37.1 \\
			& AVT+~\cite{activipative_transformer} & 34.4 & 20.2 & 16.8 & 80.0 & 51.6 & 36.5 \\
			& A-ACT~\cite{Gupta2022action} & 36.0 & 24.3 & 16.6 & 80.1 & 53.5 & 36.7 \\
			& DCR~\cite{xu2022learning}  & / & / & 17.7 & / & / & 38.5 \\
			\cmidrule{2-8}
			& RULSTM~\cite{rulstm} & 33.0 & 22.8 & 14.4 & 79.6 & 51.0 & 33.7 \\
			& \bf Ub-RULSTM & \bf 33.0 & \bf 23.0 & \bf 14.8 & \bf 79.7 & \bf 52.2 & \bf 34.5 \\
			\cmidrule{2-8}
			& DCR(TSM)~\cite{xu2022learning} & 31.0 & \bf 18.2 & 11.7 & 64.1 & 41.4 & 26.4 \\
			& \bf Ub-DCR(TSM) & \bf 33.9 & 17.9 & \bf 11.8 & \bf 75.7 & \bf 42.1 & \bf 28.2 \\
			\midrule
			\multirow{5}{*}{\textbf{S2}}
			& ATSN~\cite{damen2018scaling} & 25.3 & 10.4 & 2.4 & 68.3 & 29.5 & 6.6 \\
			& MCE~\cite{furnari2018leveraging} & 21.3 & 9.9 & 5.6 & 63.3 & 25.5 & 15.7 \\
			& ActionBanks~\cite{sener2020temporal} & 27.5 & 16.6 & 10.0 & 66.8 & 32.8 & 23.4 \\
			& SRL~\cite{rulstm} & 27.4 & 15.5 & 8.9 & 71.9 & 36.8 & 22.1 \\
			& ImagineRNN~\cite{wu2020learning} & 29.3 & 15.5 & 9.3 & 70.7 & 35.8 & 22.2 \\
			& MMTCN-TBN~\cite{zatsarynna2021multi} & 30.7 & 14.9 & 8.9 & 72.0 & 36.7 & 21.7 \\
			& FHOI~\cite{liu2020forecasting} & 28.3 & 14.1 & 8.6 & 70.7 & 34.4 & 22.9 \\
			& Ego-OMG~\cite{dessalene2021forecasting} & 27.4 & 17.7 & 11.8 & 68.6 & 37.9 & 23.8 \\
			& MGRKD~\cite{huang20multimodal} & 29.3 & 16.6 & 10.4 & 70.8 & 37.8 & 23.1 \\
			& A-ACT~\cite{Gupta2022action} & 29.2 & 16.0 & 10.3 & 71.1 & 36.5 & 23.5 \\
			& AVT+~\cite{activipative_transformer} & 30.7 & 15.6 & 10.4 & 72.2 & 40.8 & 24.3 \\
			& DCR~\cite{xu2022learning}  & / & / & 10.9 & / & / & 24.8 \\
			\cmidrule{2-8}
			& RULSTM~\cite{rulstm} & 27.0 & 15.2 & 8.2 & 69.6 & 34.4 & 21.1 \\
			& \bf Ub-RULSTM & \bf 27.5 & \bf 15.2 & \bf 9.1 & \bf 70.0 & \bf 35.7 &  \bf21.3\\
			\cmidrule{2-8}
			& DCR(TSM)~\cite{xu2022learning} & 25.1 & 9.5 & 5.4 & 52.8 & 23.4 & 13.6 \\
			& \bf Ub-DCR(TSM) & \bf 25.6 & \bf 9.8 & \bf 5.9 & \bf 65.8 & \bf 26.1 & \bf 14.5 \\
			\bottomrule
		\end{tabular}
  \vspace{-2mm}
	\label{table_epic55_test}
\end{table}

\begin{table}[t]
	\centering
	\renewcommand{\arraystretch}{0.95}
	\setlength{\tabcolsep}{1.4mm}
	\caption{Ablation studies on the number of samples in the sample-wise relative uncertainty learning.}
   \vspace{-2mm}
		\begin{tabular}{c c c c c c c c c}
			\toprule
			\multirow{2}{*}{Number} & \multicolumn{8}{c}{Top-5 Accuracy \% at different $\tau_a$ (s)} \\
			\cmidrule{2-9}
			& 2 & 1.75 & 1.5 & 1.25 & 1.0 & 0.75 & 0.5 & 0.25 \\
			\midrule
			2 & \bf 27.1 & \bf 28.5 & \bf 29.2 & \bf 30.5 & \bf 32.1 & \bf 33.7 & \bf 34.7 & \bf 36.1 \\
			3 & 26.6 & 27.8 & 28.7 & 29.6 & 31.6 & 32.5 & 34.3 & 35.8 \\
			5 & 25.2 & 27.1 & 28.0 & 29.0 & 30.6 & 31.5 & 33.1 & 35.4 \\
			10 & 24.3 & 24.8 & 25.7 & 26.8 & 28.2 & 29.3 & 30.8 & 32.3 \\
			\bottomrule
		\end{tabular}
  \vspace{-4mm}
	\label{table_epic55_ablation_mixupnumber}
\end{table}

\textbf{The Effect of the Sample-wise Relative Uncertainty Learning.} 
As shown in~\Cref{table_epic55_ablation_rgb}, without the sample-wise relative uncertainty learning, the anticipation performance is lower than Ub-Baseline. 
Besides, we conduct experiments to see the effect of the number of samples employed in this strategy. As shown in~\Cref{table_epic55_ablation_mixupnumber}, utilizing more samples will hinder the validity of this strategy and result in poor anticipation performance at all timestamps, because of the reduced diversity on the relative uncertainty value. 
With the increasing number of samples, the difference between their normalized relative uncertainty value will gradually be narrowed, while the original absolute uncertainty value differs greatly. 
Similar relative uncertainty values can not guarantee the mixed feature to contain more information from hard samples. 
Using such over-smoothed uncertainty values to adjust the probability distribution of target activity categories will lead to poor performance.

\textbf{The Effect of the Temporal Relative Uncertainty Learning.}
From~\Cref{table_epic55_ablation_rgb}, we can find that the anticipation performance is degraded at varying degrees at most timestamps without the temporal relative uncertainty learning strategy, which shows that this strategy is particularly useful in modeling video content with temporal evolution characteristics.

\textbf{The Effect of the Mean Operation on Uncertainty Vector.} 
We compare the mean pooling operation we utilized with maximum and minimum pooling operations to assess the effectiveness of the way to obtain the uncertainty value. 
As shown in Table~\ref{table_epic55_ablation_meanvalue}, the results indicate that the mean pooling operation achieves the best anticipation performance. The superiority of mean pooling stems from its inclusion of all elements in the uncertainty vector, providing an overall understanding of data uncertainty. It results in a more precise adjustment of the probability distribution for target activities based on video data and activity evolution characteristics. Furthermore, the mean operation reduces the impact of noise and outliers within the uncertainty vector. As a result, it yields a more stable estimate that is less susceptible to extreme values.

\begin{table}[t]
	\centering
	\renewcommand{\arraystretch}{0.95}
	\setlength{\tabcolsep}{1.4mm}
	\caption{Ablation studies about the mean operation on uncertainty vector.}
   \vspace{-2mm}
		\begin{tabular}{c c c c c c c c c}
			\toprule
			\multirow{2}{*}{Setting} & \multicolumn{8}{c}{Top-5 Accuracy \% at different $\tau_a$ (s)} \\
			\cmidrule{2-9}
			& 2 & 1.75 & 1.5 & 1.25 & 1.0 & 0.75 & 0.5 & 0.25 \\
			\midrule
			max & 26.5 & 27.4 & 28.8 & 29.8 & 31.8 & 32.9 & 33.4 & 35.6 \\
			min & 26.9 & 27.9 & 28.9 & 30.0 & 31.6 & 33.2 & 34.9 & 35.9 \\
			mean & \bf 27.1 & \bf 28.5 & \bf 29.2 & \bf 30.5 & \bf 32.1 & \bf 33.7 & \bf 34.7 & \bf 36.1 \\
			\bottomrule
		\end{tabular}
    \vspace{-2mm}
	\label{table_epic55_ablation_meanvalue}
\end{table}

\begin{table}[t]
	\centering
	\renewcommand{\arraystretch}{0.95}
	\setlength{\tabcolsep}{1.4mm}
	\caption{Ablation studies on the internal and external uncertainty matrices.}
 \vspace{-2mm}
		\begin{tabular}{c c c c c c c c c}
			\toprule
			\multirow{2}{*}{Setting} & \multicolumn{8}{c}{Top-5 Accuracy \% at different $\tau_a$ (s)} \\
			\cmidrule{2-9}
			& 2 & 1.75 & 1.5 & 1.25 & 1.0 & 0.75 & 0.5 & 0.25 \\
			\midrule
			Baseline & 26.2 & 26.8 & 28.2 & 29.7 & 30.9 & 32.1 & 33.3 & 34.0 \\
			\midrule
			+Ex & 26.4 & 27.2 & 28.5 & 30.1 & 31.5 & 32.5 & 33.9 & 35.3 \\
			+In & 26.9 & 28.1 & 29.2 & 30.3 & 31.8 & 32.7 & 34.1 & 35.4 \\
			+Ex \& In & \bf 27.1 & \bf 28.5 & \bf 29.2 & \bf 30.5 & \bf 32.1 & \bf 33.7 & \bf 34.7 & \bf 36.1 \\
			\bottomrule
		\end{tabular}
  \vspace{-4mm}
	\label{table_epic55_ablation_scoingfunction}
\end{table}

\textbf{The Effect of the Constructed Labels.}
By comparing the Ub-Baseline and `-optlabel' in~\Cref{table_epic55_ablation_rgb}, it becomes evident that our constructed target activity labels consistently improve the anticipation performance in most cases, which proves the validity of this strategy. Further, we reveal the advantages of this strategy from two aspects.

First, we evaluate the rationality of the internal and external uncertainty matrices used in this strategy. We introduce three variants of this strategy to train the Baseline model. The `Ex' means only using the external uncertainty matrix, the `In' indicates only using the internal uncertainty matrix, and the `Ex \& In' means using both. As presented in~\Cref{table_epic55_ablation_scoingfunction}, the external uncertainty matrix boosts the top-5 accuracy of the anticipation results from 34.0\% to 35.3\% at anticipation timestamp 0.25s. 
Additionally, we can well perceive the effectiveness of the internal uncertainty matrix by observing the performance gap between Baseline and Baseline+In. The performance is further improved when we use both uncertainty matrices. Notably, the performance gap between Baseline and the `In' is more significant than that between Baseline and the `Ex', which indicates that the internal uncertainty matrix provides more valuable guidance than the external uncertainty matrix since it is directly obtained through dataset statistics. In comparison, the knowledge expressed by the external uncertainty matrix tends to be more general and dataset-independent.

\begin{table*}[t]
	\centering
	\renewcommand{\arraystretch}{0.95}
	\setlength{\tabcolsep}{1.1mm}
	\caption{Ablation study on the generalization ability of the relative uncertainty learning strategy.}
 \vspace{-2mm}
		\begin{tabular}{c c c c c c c c c c c c c c c c}
			\toprule
			\multirow{2}{*}{Model} & \multicolumn{3}{c}{Overall \% @ 1s} & \multicolumn{3}{c}{Unseen \% @ 1s} & \multicolumn{3}{c}{Tail \% @ 1s} & \multicolumn{3}{c}{Uncertain \% @ 1s} & \multicolumn{3}{c}{Tail \& Uncertain \% @ 1s} \\
			\cmidrule{2-16}
			& Verb & Noun & Act & Verb & Noun & Act & Verb & Noun & Act & Verb & Noun & Act & Verb & Noun & Act \\
			\midrule
			Ub-DCR* & 28.7 & 29.7 & 12.9 & 29.3 & 24.9 & 10.9 & 22.9 & 23.5 & 11.0 & 27.8 & 40.2 & 15.7 & 21.5 & 28.1 & 12.7 \\
			\bf Ub-DCR & \bf 32.8 & \bf 34.1 & \bf 16.2 & \bf 33.0 & \bf 25.0 & \bf 11.6 & \bf 28.9 & \bf 30.4 & \bf 15.7 & \bf 32.2 & \bf 40.9 & \bf 16.4 & \bf 27.7 & \bf 32.0 & \bf 14.7 \\
			\bottomrule
		\end{tabular}
  \vspace{-2mm}
	\label{table_epic100_ablation_rul}
\end{table*}

\begin{table}[t]
	\centering
	\renewcommand{\arraystretch}{0.95}
	\setlength{\tabcolsep}{1.4mm}
	\caption{Ablation studies on the strength of the proposed labels.}
 \vspace{-2mm}
		\begin{tabular}{c c c c c c c c c}
			\toprule
			\multirow{2}{*}{Setting} & \multicolumn{8}{c}{Top-5 Accuracy \% at different $\tau_a$ (s)} \\
			\cmidrule{2-9}
			& 2 & 1.75 & 1.5 & 1.25 & 1.0 & 0.75 & 0.5 & 0.25 \\
			\midrule
			one-hot & 26.0 & 26.9 & 27.8 & 29.4 & 31.2 & 32.0 & 33.5 & 35.0 \\
			ls (0.2) & 27.0 & 27.6 & 29.1 & 30.1 & 31.8 & 33.1 & 34.5 & 35.8 \\
			ls (0.4) & 27.0 & 27.8 & 29.2 & 30.3 & 32.0 & 32.9 & 34.2 & 35.8 \\
			ours & \bf 27.1 & \bf 28.5 & \bf 29.2 & \bf 30.5 & \bf 32.1 & \bf 33.7 & \bf 34.7 & \bf 36.1 \\
			\bottomrule
		\end{tabular}
  \vspace{-4mm}
	\label{table_epic55_ablation_labeltype}
\end{table}

Second, we validate the strength of our proposed label representation compared to one-hot labels and the label smoothing approach. We construct four kinds of labels to train the Ub-Baseline, `one-hot' means using the original target activity label, `ls (0.2)' or `ls (0.4)' means using label smoothing with smooth value 0.2 or 0.4. As shown in~\Cref{table_epic55_ablation_labeltype}, the label smoothing strategy can improve the anticipation performance because it can avoid model over-fitting. Our strategy surpasses them all by better treatment on the correlated activity labels in the video content evolution.

\textbf{The Generalization Ability of the Relative Uncertainty Learning Strategy.} 		
To demonstrate the generalization ability of our relative uncertainty learning strategy when confronted with high uncertainty categories in long-tailed distributions setting, we construct the `Uncertain' Classes and `Tail \& Uncertain' Classes validation subsets on EK100 except the original Unseen Classes and Tail Classes validation subsets. 
In particular, for each activity class pair, we merge the acquired internal and external uncertainty values and then rank them by their merged values in descending order. Subsequently, we select activity classes whose merged value is greater than the given threshold to form the `Uncertain' Classes validation subset. Correspondingly, the `Tail \& Uncertain' Classes validation subset are activity categories shared by the subsets of tail classes and uncertain classes in the validation set. 
We compare Ub-DCR and Ub-DCR*, where Ub-DCR* is identical to Ub-DCR except for excluding the relative uncertainty learning strategy. 	
On this basis, we compare the performance gap between Ub-DCR and Ub-DCR* on `Unseen' classes, `Tail` classes, `Uncertain' classes, and `Tail \& Uncertain' classes in Table~\ref{table_epic100_ablation_rul}. We conclude that our relative uncertainty learning strategy significantly enhances the generalization capability when handling `Unseen' classes, `Tail' classes and classes of high uncertainty.

\begin{figure}[t]
	\centering
	\subfigure[The top-5 anticipation results of the DCR and Ub-DCR.
	{\label{fig_visualization_dcr}}]
	{\includegraphics[width=1\linewidth]{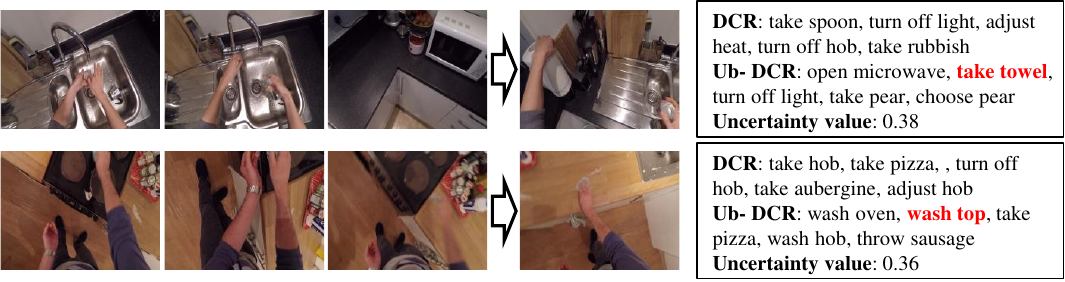}}
	\quad
	\subfigure[The top-5 anticipation results of the RULSTM and Ub-RULSTM. {\label{fig_visualization_rulstm}}]
	{\includegraphics[width=1\linewidth]{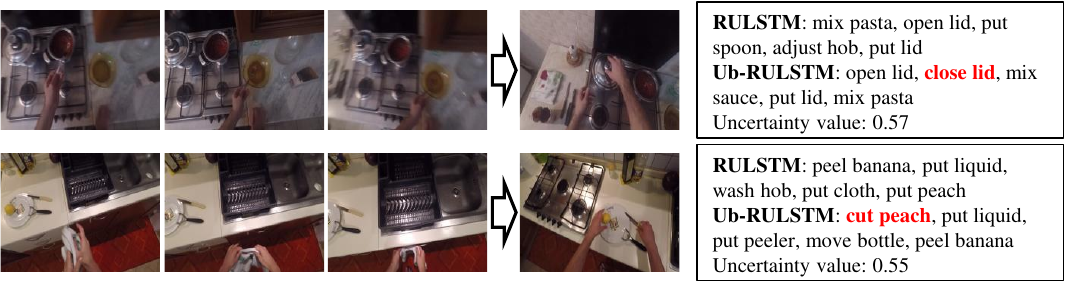}}
	\quad
 \vspace{-2mm}
	\caption{Visualization of the anticipation results. The ground-truth activity category is marked with red color.}
	\label{fig_visualization}
\end{figure}

\textbf{Qualitative Analysis}. We visualize the anticipation results of DCR, Ub-DCR, RULSTM and Ub-RULSTM backbones on the RGB feature from TSN on EK100 validation set in Figure~\ref{fig_visualization_dcr} and Figure~\ref{fig_visualization_rulstm}. We can see that there are situations in which Ub-DCR (or Ub-RULSTM) gives the correct target activity anticipation results, but DCR (or RULSTM) does not. 

\begin{figure}[t]
	\centering
	\includegraphics[scale=0.475]{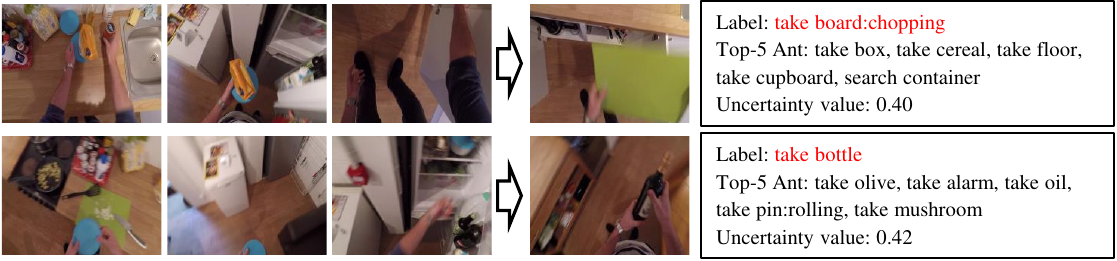}
	\caption{The failure case visualization. The top-5 results of Ub-DCR on anticipation timestamp 1s are shown on the right.}
\vspace{-4mm}
	\label{fig_visualization_failed}
\end{figure}

To enhance deeper comprehension of our framework, we present some failure cases in Figure~\ref{fig_visualization_failed}. In the first instance, a lack of significant semantic correlation between observed activities and the target activity renders our constructed labels less effective, resulting in erroneous anticipation. In the second example, given the observed video clip, despite anticipating the operator retrieving objects from the refrigerator, the observed video lacks information specific to the target activity. Given the numerous potential future activities, our framework's anticipation may be inaccurate as the dataset assigns a single target activity. Nonetheless, our framework's top-5 anticipation results remain reasonable.

\section{Weakness}
\label{sec_weakness}

Despite that our uncertainty-boosted video activity anticipation framework performs fairly well in uncertainty modeling and improves the activity anticipation performance of various backbones, there are still several issues to consider. 
\begin{itemize}
	\item We focus on investigating the data uncertainty of the video activity anticipation task. It is worth noting that there is substantial potential for future research in developing more effective methods for simultaneously modeling both data and model uncertainty. 	
	\item The internal uncertainty matrix is dataset-depend, making it hard to be transferred to other data domain. One may need to develop a more general calculation scheme if we apply similar ideas to video foundation model training. 
	\item To ensure the accuracy of uncertainty values, we need to choose appropriate values of $\alpha$ and $\beta$ for different backbones, which leads to a slight increase in the model complexity. 
\end{itemize}

\section{Conclusion}
\label{sec_conclusion}

In this paper, we systemically investigate the uncertainty learning problem for the video activity anticipation and propose an uncertainty-boosted robust activity anticipation framework that can be plugged into a wide range of existing models with slight effort. It produces an uncertainty value to indicate the credibility of the model results. Our method significantly improves the robustness of existing models when dealing with samples with high uncertainty and activity categories with high uncertainty or long-tailed distributions. 
The proposed activity evolution uncertainty measurement provides inspiration for explainable and trustable video understanding. 
Furthermore, the distribution adjustment and relative learning strategies can enlighten various video comprehension tasks such as long-term video classification. 
In future work, we aim to develop video understanding models with strong interpretation and high performance by injecting the uncertainty modeling deeper into the model inside and at a finer granularity and accordingly, achieving better regulation of the model learning behavior.

%

\ifCLASSOPTIONcompsoc
\section*{Acknowledgments}
\else
\section*{Acknowledgment}
\fi

The authors would like to thank the associate editor and the reviewers for their time and effort provided to review the manuscript. This work was supported in part by the National Key R\&D Program of China under Grant 2023YFC2508704, and in part by National Natural Science Foundation of China: U21B2038, 62236008, and 62306092.


\ifCLASSOPTIONcaptionsoff
  \newpage
\fi



\bibliographystyle{IEEEtran}
\bibliography{IEEEabrv,reference}

\begin{thebibliography}{10}
\providecommand{\url}[1]{#1}
\csname url@samestyle\endcsname
\providecommand{\newblock}{\relax}
\providecommand{\bibinfo}[2]{#2}
\providecommand{\BIBentrySTDinterwordspacing}{\spaceskip=0pt\relax}
\providecommand{\BIBentryALTinterwordstretchfactor}{4}
\providecommand{\BIBentryALTinterwordspacing}{\spaceskip=\fontdimen2\font plus
\BIBentryALTinterwordstretchfactor\fontdimen3\font minus
  \fontdimen4\font\relax}
\providecommand{\BIBforeignlanguage}[2]{{%
\expandafter\ifx\csname l@#1\endcsname\relax
\typeout{** WARNING: IEEEtran.bst: No hyphenation pattern has been}%
\typeout{** loaded for the language `#1'. Using the pattern for}%
\typeout{** the default language instead.}%
\else
\language=\csname l@#1\endcsname
\fi
#2}}
\providecommand{\BIBdecl}{\relax}
\BIBdecl

\bibitem{de2016online}
R.~De~Geest, E.~Gavves, A.~Ghodrati, Z.~Li, C.~Snoek, and T.~Tuytelaars,
  ``Online action detection,'' in \emph{European Conference on Computer
  Vision}, 2016, pp. 269--284.

\bibitem{koppula2015anticipating}
H.~S. Koppula and A.~Saxena, ``Anticipating human activities using object
  affordances for reactive robotic response,'' \emph{IEEE transactions on
  pattern analysis and machine intelligence}, vol.~38, no.~1, pp. 14--29, 2015.

\bibitem{hutchinson2021video}
M.~Hutchinson and V.~Gadepally, ``Video action understanding,'' \emph{{IEEE}
  Access}, vol.~9, pp. 134\,611--134\,637, 2021.

\bibitem{abu2018will}
Y.~Abu~Farha, A.~Richard, and J.~Gall, ``When will you do what?-anticipating
  temporal occurrences of activities,'' in \emph{Proceedings of the IEEE
  Conference on Computer Vision and Pattern Recognition}, 2018, pp. 5343--5352.

\bibitem{rulstm}
A.~Furnari and G.~M. Farinella, ``What would you expect? anticipating
  egocentric actions with rolling-unrolling lstms and modality attention,'' in
  \emph{Proceedings of the IEEE International Conference on Computer Vision},
  2019, pp. 6252--6261.

\bibitem{srl}
Z.~Qi, S.~Wang, C.~Su, L.~Su, Q.~Huang, and Q.~Tian, ``Self-regulated learning
  for egocentric video activity anticipation,'' \emph{IEEE Transactions on
  Pattern Analysis and Machine Intelligence}, vol.~45, no.~6, pp. 6715--6730,
  2023.

\bibitem{zatsarynna2021multi}
O.~Zatsarynna, Y.~Abu~Farha, and J.~Gall, ``Multi-modal temporal convolutional
  network for anticipating actions in egocentric videos,'' in \emph{Proceedings
  of the IEEE/CVF Conference on Computer Vision and Pattern Recognition}, 2021,
  pp. 2249--2258.

\bibitem{camporese2021knowledge}
G.~Camporese, P.~Coscia, A.~Furnari, G.~M. Farinella, and L.~Ballan,
  ``Knowledge distillation for action anticipation via label smoothing,'' in
  \emph{2020 25th International Conference on Pattern Recognition}, 2021, pp.
  3312--3319.

\bibitem{fernando2021anticipating}
B.~Fernando and S.~Herath, ``Anticipating human actions by correlating past
  with the future with jaccard similarity measures,'' in \emph{Proceedings of
  the IEEE/CVF Conference on Computer Vision and Pattern Recognition}, 2021,
  pp. 13\,224--13\,233.

\bibitem{qi2017predicting}
S.~Qi, S.~Huang, P.~Wei, and S.-C. Zhu, ``Predicting human activities using
  stochastic grammar,'' in \emph{Proceedings of the IEEE International
  Conference on Computer Vision}, 2017, pp. 1164--1172.

\bibitem{furnari2017next}
A.~Furnari, S.~Battiato, K.~Grauman, and G.~M. Farinella, ``Next-active-object
  prediction from egocentric videos,'' \emph{Journal of Visual Communication
  and Image Representation}, vol.~49, pp. 401--411, 2017.

\bibitem{abu2019uncertainty}
Y.~A. Farha and J.~Gall, ``Uncertainty-aware anticipation of activities,'' in
  \emph{Proceedings of the IEEE International Conference on Computer Vision
  Workshops}, 2019, pp. 1197--1204.

\bibitem{furnari2018leveraging}
A.~Furnari, S.~Battiato, and G.~M. Farinella, ``Leveraging uncertainty to
  rethink loss functions and evaluation measures for egocentric action
  anticipation,'' in \emph{European Conference on Computer Vision}, 2018, pp.
  389--405.

\bibitem{ke2019time}
Q.~Ke, M.~Fritz, and B.~Schiele, ``Time-conditioned action anticipation in one
  shot,'' in \emph{Proceedings of the IEEE Conference on Computer Vision and
  Pattern Recognition}, 2019, pp. 9925--9934.

\bibitem{blundell2015weight}
C.~Blundell, J.~Cornebise, K.~Kavukcuoglu, and D.~Wierstra, ``Weight
  uncertainty in neural network,'' in \emph{International conference on machine
  learning}, 2015, pp. 1613--1622.

\bibitem{gal2016dropout}
Y.~Gal and Z.~Ghahramani, ``Dropout as a bayesian approximation: Representing
  model uncertainty in deep learning,'' in \emph{international conference on
  machine learning}.\hskip 1em plus 0.5em minus 0.4em\relax PMLR, 2016, pp.
  1050--1059.

\bibitem{kendall2017uncertainties}
A.~Kendall and Y.~Gal, ``What uncertainties do we need in bayesian deep
  learning for computer vision?'' \emph{Advances in neural information
  processing systems}, vol.~30, 2017.

\bibitem{hora1996aleatory}
S.~C. Hora, ``Aleatory and epistemic uncertainty in probability elicitation
  with an example from hazardous waste management,'' \emph{Reliability
  Engineering \& System Safety}, vol.~54, no. 2-3, pp. 217--223, 1996.

\bibitem{der2009aleatory}
A.~Der~Kiureghian and O.~Ditlevsen, ``Aleatory or epistemic? does it matter?''
  \emph{Structural safety}, vol.~31, no.~2, pp. 105--112, 2009.

\bibitem{ayhan2022test}
M.~S. Ayhan and P.~Berens, ``Test-time data augmentation for estimation of
  heteroscedastic aleatoric uncertainty in deep neural networks,'' in
  \emph{Medical Imaging with Deep Learning}, 2022.

\bibitem{depeweg2018decomposition}
S.~Depeweg, J.-M. Hernandez-Lobato, F.~Doshi-Velez, and S.~Udluft,
  ``Decomposition of uncertainty in bayesian deep learning for efficient and
  risk-sensitive learning,'' in \emph{International Conference on Machine
  Learning}.\hskip 1em plus 0.5em minus 0.4em\relax PMLR, 2018, pp. 1184--1193.

\bibitem{hullermeier2021aleatoric}
E.~H{\"u}llermeier and W.~Waegeman, ``Aleatoric and epistemic uncertainty in
  machine learning: An introduction to concepts and methods,'' \emph{Machine
  Learning}, vol. 110, pp. 457--506, 2021.

\bibitem{gruber2023sources}
C.~Gruber, P.~O. Schenk, M.~Schierholz, F.~Kreuter, and G.~Kauermann, ``Sources
  of uncertainty in machine learning--a statisticians' view,'' \emph{arXiv
  preprint arXiv:2305.16703}, 2023.

\bibitem{paulsen2016introduction}
V.~I. Paulsen and M.~Raghupathi, \emph{An introduction to the theory of
  reproducing kernel Hilbert spaces}.\hskip 1em plus 0.5em minus 0.4em\relax
  Cambridge university press, 2016, vol. 152.

\bibitem{damen2018scaling}
D.~Damen, H.~Doughty, G.~Maria~Farinella, S.~Fidler, A.~Furnari, E.~Kazakos,
  D.~Moltisanti, J.~Munro, T.~Perrett, W.~Price \emph{et~al.}, ``Scaling
  egocentric vision: The epic-kitchens dataset,'' in \emph{Proceedings of the
  European Conference on Computer Vision}, 2018, pp. 720--736.

\bibitem{speer2017conceptnet5}
R.~Speer, J.~Chin, and C.~Havasi, ``Conceptnet 5.5: An open multilingual graph
  of general knowledge,'' in \emph{Thirty-first AAAI conference on artificial
  intelligence}, 2017.

\bibitem{EPIC-100}
D.~Damen, H.~Doughty, G.~M. Farinella, A.~Furnari, E.~Kazakos, J.~Ma,
  D.~Moltisanti, J.~Munro, T.~Perrett, W.~Price, and M.~Wray, ``Rescaling
  egocentric vision: Collection, pipeline and challenges for
  {EPIC-KITCHENS-100},'' \emph{Int. J. Comput. Vis.}, vol. 130, no.~1, pp.
  33--55, 2022.

\bibitem{li2018eye}
Y.~Li, M.~Liu, and J.~M. Rehg, ``In the eye of beholder: Joint learning of gaze
  and actions in first person video,'' in \emph{Proceedings of the European
  Conference on Computer Vision}, 2018, pp. 619--635.

\bibitem{ragusa2023meccano}
F.~Ragusa, A.~Furnari, and G.~M. Farinella, ``Meccano: A multimodal egocentric
  dataset for humans behavior understanding in the industrial-like domain,''
  \emph{Computer Vision and Image Understanding}, p. 103764, 2023.

\bibitem{stein2013combining}
S.~Stein and S.~J. McKenna, ``Combining embedded accelerometers with computer
  vision for recognizing food preparation activities,'' in \emph{Proceedings of
  the 2013 ACM international joint conference on Pervasive and ubiquitous
  computing}.\hskip 1em plus 0.5em minus 0.4em\relax ACM, 2013, pp. 729--738.

\bibitem{wu2022memvit}
C.-Y. Wu, Y.~Li, K.~Mangalam, H.~Fan, B.~Xiong, J.~Malik, and C.~Feichtenhofer,
  ``Memvit: Memory-augmented multiscale vision transformer for efficient
  long-term video recognition,'' in \emph{Proceedings of the IEEE/CVF
  Conference on Computer Vision and Pattern Recognition}, 2022, pp.
  13\,587--13\,597.

\bibitem{babaeizadeh2018stochastic}
M.~Babaeizadeh, C.~Finn, D.~Erhan, R.~H. Campbell, and S.~Levine, ``Stochastic
  variational video prediction,'' in \emph{International Conference on Learning
  Representations}, 2018.

\bibitem{rudenko2020human}
A.~Rudenko, L.~Palmieri, M.~Herman, K.~M. Kitani, D.~M. Gavrila, and K.~O.
  Arras, ``Human motion trajectory prediction: A survey,'' \emph{The
  International Journal of Robotics Research}, vol.~39, no.~8, pp. 895--935,
  2020.

\bibitem{wang2021self}
J.~Wang, J.~Jiao, L.~Bao, S.~He, W.~Liu, and Y.-H. Liu, ``Self-supervised video
  representation learning by uncovering spatio-temporal statistics,''
  \emph{IEEE Transactions on Pattern Analysis and Machine Intelligence},
  vol.~44, no.~7, pp. 3791--3806, 2021.

\bibitem{wu2017anticipating}
T.-Y. Wu, T.-A. Chien, C.-S. Chan, C.-W. Hu, and M.~Sun, ``Anticipating daily
  intention using on-wrist motion triggered sensing,'' in \emph{Proceedings of
  the IEEE International Conference on Computer Vision}, 2017, pp. 48--56.

\bibitem{fan2018forecasting}
C.~Fan, J.~Lee, and M.~S. Ryoo, ``Forecasting hands and objects in future
  frames,'' in \emph{European Conference on Computer Vision}, 2018, pp.
  124--137.

\bibitem{rhinehart2017first}
N.~Rhinehart and K.~M. Kitani, ``First-person activity forecasting with online
  inverse reinforcement learning,'' in \emph{Proceedings of the IEEE
  International Conference on Computer Vision}, 2017, pp. 3696--3705.

\bibitem{zhang2017deep}
M.~Zhang, K.~Teck~Ma, J.~Hwee~Lim, Q.~Zhao, and J.~Feng, ``Deep future gaze:
  Gaze anticipation on egocentric videos using adversarial networks,'' in
  \emph{Proceedings of the IEEE conference on computer vision and pattern
  recognition}, 2017, pp. 4372--4381.

\bibitem{zhang2020egocentric}
T.~Zhang, W.~Min, Y.~Zhu, Y.~Rui, and S.~Jiang, ``An egocentric action
  anticipation framework via fusing intuition and analysis,'' in
  \emph{Proceedings of the 28th ACM International Conference on Multimedia},
  2020, pp. 402--410.

\bibitem{felsen2017will}
P.~Felsen, P.~Agrawal, and J.~Malik, ``What will happen next? forecasting
  player moves in sports videos,'' in \emph{Proceedings of the IEEE
  International Conference on Computer Vision}, 2017, pp. 3342--3351.

\bibitem{zhao2020diverse}
H.~Zhao and R.~P. Wildes, ``On diverse asynchronous activity anticipation,'' in
  \emph{Proceedings of the European Conference on Computer Vision}, 2020, pp.
  781--799.

\bibitem{zhang20anegocentric}
T.~Zhang, W.~Min, Y.~Zhu, Y.~Rui, and S.~Jiang, ``An egocentric action
  anticipation framework via fusing intuition and analysis,'' in
  \emph{Proceedings of the ACM International Conference on Multimedia}, 2020,
  pp. 402--410.

\bibitem{nawhal2022rethinking}
M.~Nawhal, A.~A. Jyothi, and G.~Mori, ``Rethinking learning approaches for
  long-term action anticipation,'' in \emph{European Conference on Computer
  Vision}, 2022, pp. 558--576.

\bibitem{zhang2024object}
C.~Zhang, C.~Fu, S.~Wang, N.~Agarwal, K.~Lee, C.~Choi, and C.~Sun,
  ``Object-centric video representation for long-term action anticipation,'' in
  \emph{Proceedings of the IEEE/CVF Winter Conference on Applications of
  Computer Vision}, 2024, pp. 6751--6761.

\bibitem{wang2023memory}
J.~Wang, G.~Chen, Y.~Huang, L.~Wang, and T.~Lu, ``Memory-and-anticipation
  transformer for online action understanding,'' in \emph{Proceedings of the
  IEEE/CVF International Conference on Computer Vision}, 2023, pp.
  13\,824--13\,835.

\bibitem{girase2023latency}
H.~Girase, N.~Agarwal, C.~Choi, and K.~Mangalam, ``Latency matters: Real-time
  action forecasting transformer,'' in \emph{Proceedings of the IEEE/CVF
  Conference on Computer Vision and Pattern Recognition}, 2023, pp.
  18\,759--18\,769.

\bibitem{thakur2024leveraging}
S.~Thakur, C.~Beyan, P.~Morerio, V.~Murino, and A.~Del~Bue, ``Leveraging
  next-active objects for context-aware anticipation in egocentric videos,'' in
  \emph{Proceedings of the IEEE/CVF Winter Conference on Applications of
  Computer Vision}, 2024, pp. 8657--8666.

\bibitem{activipative_transformer}
R.~Girdhar and K.~Grauman, ``Anticipative video transformer,'' in
  \emph{Proceedings of the IEEE/CVF International Conference on Computer
  Vision}, 2021, pp. 13\,505--13\,515.

\bibitem{zhong2023diffant}
Z.~Zhong, C.~Wu, M.~Martin, M.~Voit, J.~Gall, and J.~Beyerer, ``Diffant:
  Diffusion models for action anticipation,'' \emph{arXiv preprint
  arXiv:2311.15991}, 2023.

\bibitem{yu2023merlin}
E.~Yu, L.~Zhao, Y.~Wei, J.~Yang, D.~Wu, L.~Kong, H.~Wei, T.~Wang, Z.~Ge,
  X.~Zhang \emph{et~al.}, ``Merlin: Empowering multimodal llms with foresight
  minds,'' \emph{arXiv preprint arXiv:2312.00589}, 2023.

\bibitem{zhao2023antgpt}
Q.~Zhao, C.~Zhang, S.~Wang, C.~Fu, N.~Agarwal, K.~Lee, and C.~Sun, ``Antgpt:
  Can large language models help long-term action anticipation from videos?''
  \emph{arXiv preprint arXiv:2307.16368}, 2023.

\bibitem{kim2023lalm}
S.~Kim, D.~Huang, Y.~Xian, O.~Hilliges, L.~Van~Gool, and X.~Wang, ``Lalm:
  Long-term action anticipation with language models,'' \emph{arXiv preprint
  arXiv:2311.17944}, 2023.

\bibitem{sadegh2017encouraging}
M.~Sadegh~Aliakbarian, F.~Sadat~Saleh, M.~Salzmann, B.~Fernando, L.~Petersson,
  and L.~Andersson, ``Encouraging lstms to anticipate actions very early,'' in
  \emph{Proceedings of the IEEE International Conference on Computer Vision},
  2017, pp. 280--289.

\bibitem{mahmud2017joint}
T.~Mahmud, M.~Hasan, and A.~K. Roy-Chowdhury, ``Joint prediction of activity
  labels and starting times in untrimmed videos,'' in \emph{Proceedings of the
  IEEE International Conference on Computer Vision}, 2017, pp. 5773--5782.

\bibitem{xiao2019quantifying}
Y.~Xiao and W.~Y. Wang, ``Quantifying uncertainties in natural language
  processing tasks,'' in \emph{Proceedings of the AAAI Conference on Artificial
  Intelligence}, vol.~33, no.~01, 2019, pp. 7322--7329.

\bibitem{gal2016theoretically}
Y.~Gal and Z.~Ghahramani, ``A theoretically grounded application of dropout in
  recurrent neural networks,'' \emph{Advances in neural information processing
  systems}, vol.~29, 2016.

\bibitem{zhu2017deep}
L.~Zhu and N.~Laptev, ``Deep and confident prediction for time series at
  uber,'' in \emph{2017 IEEE International Conference on Data Mining
  Workshops}, 2017, pp. 103--110.

\bibitem{kendall2015bayesian}
A.~Kendall, V.~Badrinarayanan, and R.~Cipolla, ``Bayesian segnet: Model
  uncertainty in deep convolutional encoder-decoder architectures for scene
  understanding,'' in \emph{British Machine Vision Conference}, 2017.

\bibitem{isobe2017deep}
S.~Isobe and S.~Arai, ``Deep convolutional encoder-decoder network with model
  uncertainty for semantic segmentation,'' in \emph{2017 IEEE International
  Conference on INnovations in Intelligent SysTems and Applications}, 2017, pp.
  365--370.

\bibitem{choi2019gaussian}
J.~Choi, D.~Chun, H.~Kim, and H.-J. Lee, ``Gaussian yolov3: An accurate and
  fast object detector using localization uncertainty for autonomous driving,''
  in \emph{Proceedings of the IEEE/CVF International Conference on Computer
  Vision}, 2019, pp. 502--511.

\bibitem{yu2019robust}
T.~Yu, D.~Li, Y.~Yang, T.~M. Hospedales, and T.~Xiang, ``Robust person
  re-identification by modelling feature uncertainty,'' in \emph{Proceedings of
  the IEEE/CVF International Conference on Computer Vision}, 2019, pp.
  552--561.

\bibitem{chang2020data}
J.~Chang, Z.~Lan, C.~Cheng, and Y.~Wei, ``Data uncertainty learning in face
  recognition,'' in \emph{Proceedings of the IEEE/CVF Conference on Computer
  Vision and Pattern Recognition}, 2020, pp. 5710--5719.

\bibitem{zhang2021relative}
Y.~Zhang, C.~Wang, and W.~Deng, ``Relative uncertainty learning for facial
  expression recognition,'' \emph{Advances in Neural Information Processing
  Systems}, vol.~34, pp. 17\,616--17\,627, 2021.

\bibitem{chen2020monopair}
Y.~Chen, L.~Tai, K.~Sun, and M.~Li, ``Monopair: Monocular 3d object detection
  using pairwise spatial relationships,'' in \emph{Proceedings of the IEEE/CVF
  Conference on Computer Vision and Pattern Recognition}, 2020, pp.
  12\,093--12\,102.

\bibitem{zhou2021model}
L.~Zhou, A.~Ledent, Q.~Hu, T.~Liu, J.~Zhang, and M.~Kloft, ``Model uncertainty
  guides visual object tracking,'' in \emph{Proceedings of the AAAI Conference
  on Artificial Intelligence}, vol.~35, no.~4, 2021, pp. 3581--3589.

\bibitem{feng2018towards}
D.~Feng, L.~Rosenbaum, and K.~Dietmayer, ``Towards safe autonomous driving:
  Capture uncertainty in the deep neural network for lidar 3d vehicle
  detection,'' in \emph{2018 21st international conference on intelligent
  transportation systems}, 2018, pp. 3266--3273.

\bibitem{confalonieri2021historical}
R.~Confalonieri, L.~Coba, B.~Wagner, and T.~R. Besold, ``A historical
  perspective of explainable artificial intelligence,'' \emph{Wiley
  Interdisciplinary Reviews: Data Mining and Knowledge Discovery}, vol.~11,
  no.~1, p. e1391, 2021.

\bibitem{longo2020explainable}
L.~Longo, R.~Goebel, F.~Lecue, P.~Kieseberg, and A.~Holzinger, ``Explainable
  artificial intelligence: Concepts, applications, research challenges and
  visions,'' in \emph{International Cross-Domain Conference for Machine
  Learning and Knowledge Extraction}, 2020, pp. 1--16.

\bibitem{zhi2021mgsampler}
Y.~Zhi, Z.~Tong, L.~Wang, and G.~Wu, ``Mgsampler: An explainable sampling
  strategy for video action recognition,'' in \emph{Proceedings of the IEEE/CVF
  International Conference on Computer Vision}, 2021, pp. 1513--1522.

\bibitem{szymanowicz2021x}
S.~Szymanowicz, J.~Charles, and R.~Cipolla, ``X-man: Explaining multiple
  sources of anomalies in video,'' in \emph{Proceedings of the IEEE/CVF
  Conference on Computer Vision and Pattern Recognition}, 2021, pp. 3224--3232.

\bibitem{zhuo2019explainable}
T.~Zhuo, Z.~Cheng, P.~Zhang, Y.~Wong, and M.~Kankanhalli, ``Explainable video
  action reasoning via prior knowledge and state transitions,'' in
  \emph{Proceedings of the 27th acm international conference on multimedia},
  2019, pp. 521--529.

\bibitem{qi2020modeling}
Z.~Qi, S.~Wang, C.~Su, L.~Su, W.~Zhang, and Q.~Huang, ``Modeling temporal
  concept receptive field dynamically for untrimmed video analysis,'' in
  \emph{Proceedings of the 28th ACM International Conference on Multimedia},
  2020, pp. 3798--3806.

\bibitem{qi2020towards}
Z.~Qi, S.~Wang, C.~Su, L.~Su, Q.~Huang, and Q.~Tian, ``Towards more
  explainability: concept knowledge mining network for event recognition,'' in
  \emph{Proceedings of the 28th ACM International Conference on Multimedia},
  2020, pp. 3857--3865.

\bibitem{xu2022learning}
X.~Xu, Y.-L. Li, and C.~Lu, ``Learning to anticipate future with dynamic
  context removal,'' in \emph{Proceedings of the IEEE/CVF Conference on
  Computer Vision and Pattern Recognition}, 2022, pp. 12\,734--12\,744.

\bibitem{wang2020suppressing}
K.~Wang, X.~Peng, J.~Yang, S.~Lu, and Y.~Qiao, ``Suppressing uncertainties for
  large-scale facial expression recognition,'' in \emph{Proceedings of the
  IEEE/CVF conference on computer vision and pattern recognition}, 2020, pp.
  6897--6906.

\bibitem{xia2008listwise}
F.~Xia, T.-Y. Liu, J.~Wang, W.~Zhang, and H.~Li, ``Listwise approach to
  learning to rank: theory and algorithm,'' in \emph{Proceedings of the 25th
  international conference on Machine learning}, 2008, pp. 1192--1199.

\bibitem{sener2020temporal}
F.~Sener, D.~Singhania, and A.~Yao, ``Temporal aggregate representations for
  long-range video understanding,'' in \emph{European Conference on Computer
  Vision}, 2020, pp. 154--171.

\bibitem{gong2022future}
D.~Gong, J.~Lee, M.~Kim, S.~J. Ha, and M.~Cho, ``Future transformer for
  long-term action anticipation,'' in \emph{Proceedings of the IEEE/CVF
  Conference on Computer Vision and Pattern Recognition}, 2022, pp. 3052--3061.

\bibitem{wang2016temporal}
L.~Wang, Y.~Xiong, Z.~Wang, Y.~Qiao, D.~Lin, X.~Tang, and L.~Van~Gool,
  ``Temporal segment networks: Towards good practices for deep action
  recognition,'' in \emph{European conference on computer vision}, 2016, pp.
  20--36.

\bibitem{girshick2015fast}
R.~Girshick, ``Fast r-cnn,'' in \emph{Proceedings of the IEEE international
  conference on computer vision}, 2015, pp. 1440--1448.

\bibitem{lin2019tsm}
J.~Lin, C.~Gan, and S.~Han, ``{TSM:} temporal shift module for efficient video
  understanding,'' in \emph{{IEEE/CVF} International Conference on Computer
  Vision}, 2019, pp. 7082--7092.

\bibitem{tran2019video}
D.~Tran, H.~Wang, L.~Torresani, and M.~Feiszli, ``Video classification with
  channel-separated convolutional networks,'' in \emph{Proceedings of the
  IEEE/CVF International Conference on Computer Vision}, 2019, pp. 5552--5561.

\bibitem{terhorst2020ser}
P.~Terhorst, J.~N. Kolf, N.~Damer, F.~Kirchbuchner, and A.~Kuijper, ``Ser-fiq:
  Unsupervised estimation of face image quality based on stochastic embedding
  robustness,'' in \emph{Proceedings of the IEEE/CVF conference on computer
  vision and pattern recognition}, 2020, pp. 5651--5660.

\bibitem{cui2019class}
Y.~Cui, M.~Jia, T.-Y. Lin, Y.~Song, and S.~Belongie, ``Class-balanced loss
  based on effective number of samples,'' in \emph{Proceedings of the IEEE/CVF
  conference on computer vision and pattern recognition}, 2019, pp. 9268--9277.

\bibitem{vondrick2016anticipating}
C.~Vondrick, H.~Pirsiavash, and A.~Torralba, ``Anticipating visual
  representations from unlabeled video,'' in \emph{Proceedings of the IEEE
  Conference on Computer Vision and Pattern Recognition}, 2016, pp. 98--106.

\bibitem{gao2017red}
J.~Gao, Z.~Yang, and R.~Nevatia, ``{RED:} reinforced encoder-decoder networks
  for action anticipation,'' in \emph{British Machine Vision Conference}, 2017.

\bibitem{de2018modeling}
R.~De~Geest and T.~Tuytelaars, ``Modeling temporal structure with lstm for
  online action detection,'' in \emph{2018 IEEE Winter Conference on
  Applications of Computer Vision}, 2018, pp. 1549--1557.

\bibitem{ma2016learning}
S.~Ma, L.~Sigal, and S.~Sclaroff, ``Learning activity progression in lstms for
  activity detection and early detection,'' in \emph{Proceedings of the IEEE
  Conference on Computer Vision and Pattern Recognition}, 2016, pp. 1942--1950.

\bibitem{jain2016recurrent}
A.~Jain, A.~Singh, H.~S. Koppula, S.~Soh, and A.~Saxena, ``Recurrent neural
  networks for driver activity anticipation via sensory-fusion architecture,''
  in \emph{2016 IEEE International Conference on Robotics and Automation},
  2016, pp. 3118--3125.

\bibitem{wu2020learning}
Y.~Wu, L.~Zhu, X.~Wang, Y.~Yang, and F.~Wu, ``Learning to anticipate egocentric
  actions by imagination,'' \emph{IEEE Transactions on Image Processing},
  vol.~30, pp. 1143--1152, 2020.

\bibitem{huang20multimodal}
Y.~Huang, X.~Yang, and C.~Xu, ``Multimodal global relation knowledge
  distillation for egocentric action anticipation,'' in \emph{Proceedings of
  the ACM International Conference on Multimedia}, 2021, pp. 245--254.

\bibitem{abu2021long}
Y.~Abu~Farha, Q.~Ke, B.~Schiele, and J.~Gall, ``Long-term anticipation of
  activities with cycle consistency,'' in \emph{Pattern Recognition: 42nd DAGM
  German Conference}, 2021, pp. 159--173.

\bibitem{Gupta2022action}
A.~Gupta, J.~Liu, L.~Bo, A.~K. Roy{-}Chowdhury, and T.~Mei, ``{A-ACT:} action
  anticipation through cycle transformations,'' \emph{arXiv preprint
  arXiv:2204.00942}, 2022.

\bibitem{liu2020forecasting}
M.~Liu, S.~Tang, Y.~Li, and J.~M. Rehg, ``Forecasting human-object interaction:
  joint prediction of motor attention and actions in first person video,'' in
  \emph{European Conference on Computer Vision}, 2020, pp. 704--721.

\bibitem{tai2021higher}
T.-M. Tai, G.~Fiameni, C.-K. Lee, and O.~Lanz, ``Higher order recurrent
  space-time transformer for video action prediction,'' \emph{arXiv preprint
  arXiv:2104.08665}, 2021.

\bibitem{dessalene2021forecasting}
E.~Dessalene, C.~Devaraj, M.~Maynord, C.~Ferm{\"u}ller, and Y.~Aloimonos,
  ``Forecasting action through contact representations from first person
  video,'' \emph{IEEE Transactions on Pattern Analysis and Machine
  Intelligence}, vol.~45, no.~6, pp. 6703--6714, 2021.

\end{thebibliography}

%

\begin{IEEEbiography}[{\includegraphics[width=1in,height=1.25in,clip,keepaspectratio]{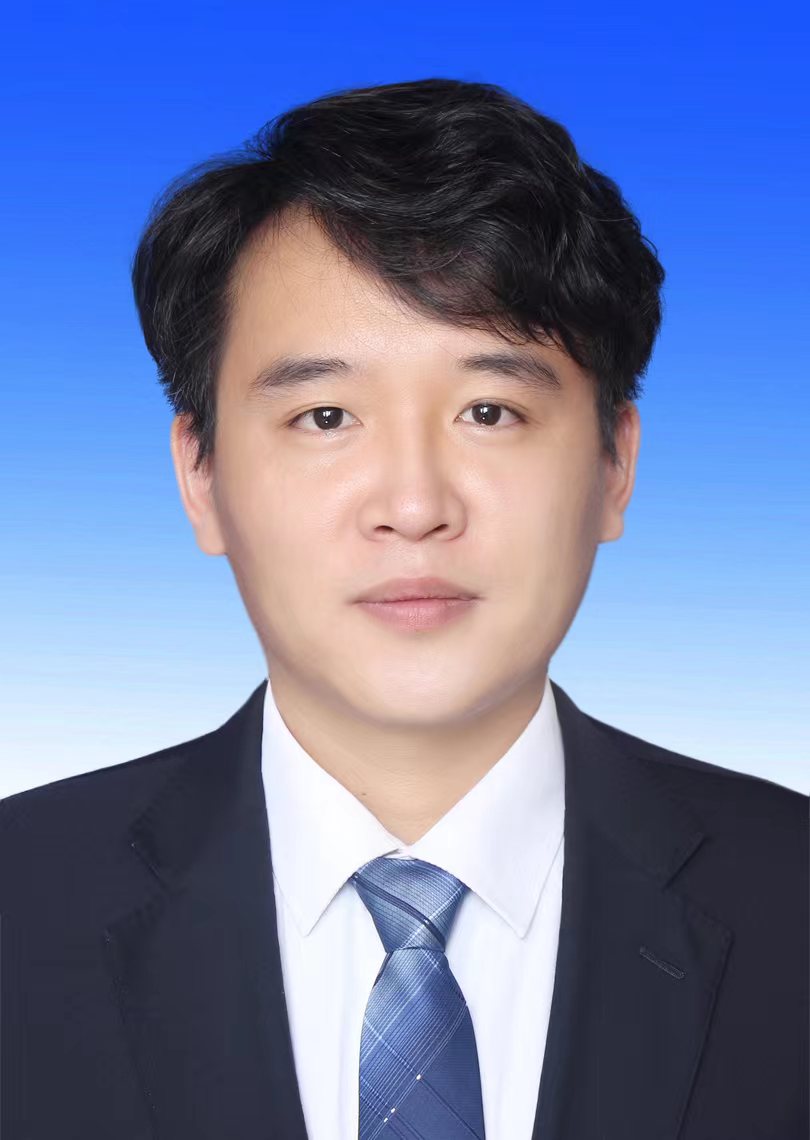}}]{Zhaobo Qi} received the B.S. degree from Harbin Institute of Technology, Weihai, China, in 2016, and the Ph.D. degree from the School of Computer Science and Technology, University of Chinese Academy of Sciences, Beijing, China, in 2022. He is currently an assistant professor in Harbin Institute of Technology, Weihai. His current research interests include video understanding, knowledge engineering and computer vision.
\end{IEEEbiography}

\begin{IEEEbiography}[{\includegraphics[width=1in,height=1.25in,clip,keepaspectratio]{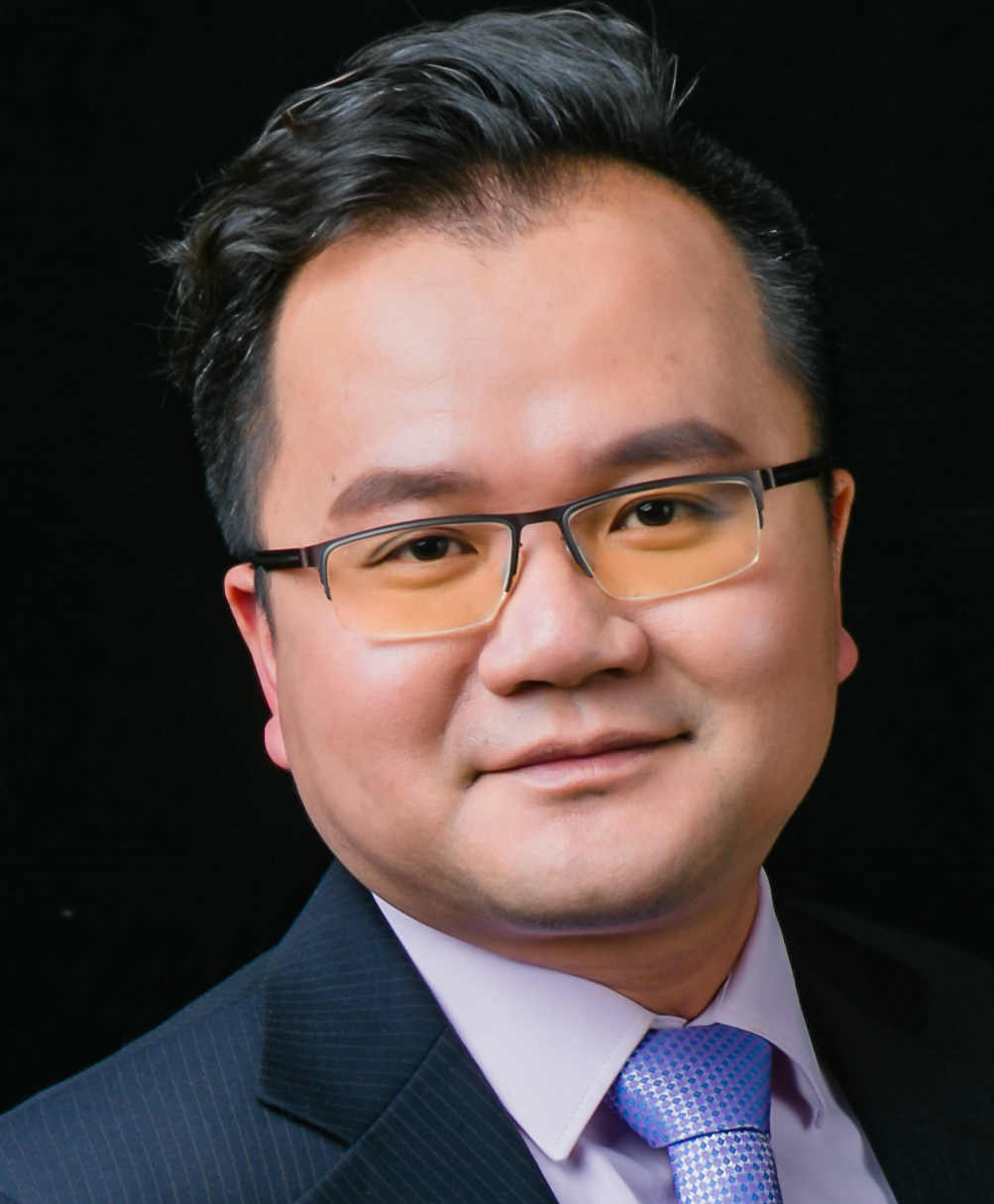}}]{Shuhui Wang} received the B.S. degree in electronics engineering from Tsinghua University, Beijing, China, in 2006, and the Ph.D. degree from the Institute of Computing Technology, Chinese Academy of Sciences, Beijing, China, in 2012. He is currently a Full Professor with the Institute of Computing Technology, Chinese Academy of Sciences. He is also with the Key Laboratory of Intelligent Information Processing, Chinese Academy of Sciences. His research interests include image/video understanding/retrieval, cross-media analysis and visual-textual knowledge extraction.
\end{IEEEbiography}

\begin{IEEEbiography}[{\includegraphics[width=1in,height=1.25in,clip,keepaspectratio]{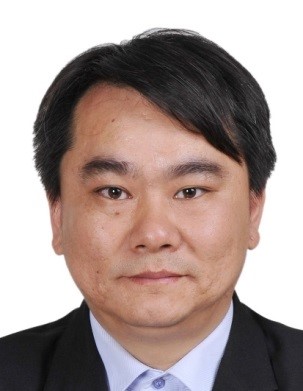}}]{Weigang Zhang } is a professor in Harbin Institute of Technology, Weihai. He received the Bachelor degree in Computer Science and Technology in 2003, the M.S. and Ph.D. degree in Computer Applied Technology in 2005 and 2016, respectively, all from Harbin Institute of Technology, China. His research interests include multimedia analysis, computer vision and pattern recognition. He has published more than 70 academic papers and is the recipient of the Best Student Paper Award at IEEE MIPR 2018.
\end{IEEEbiography}

\begin{IEEEbiography}[{\includegraphics[width=1in,height=1.25in,clip,keepaspectratio]{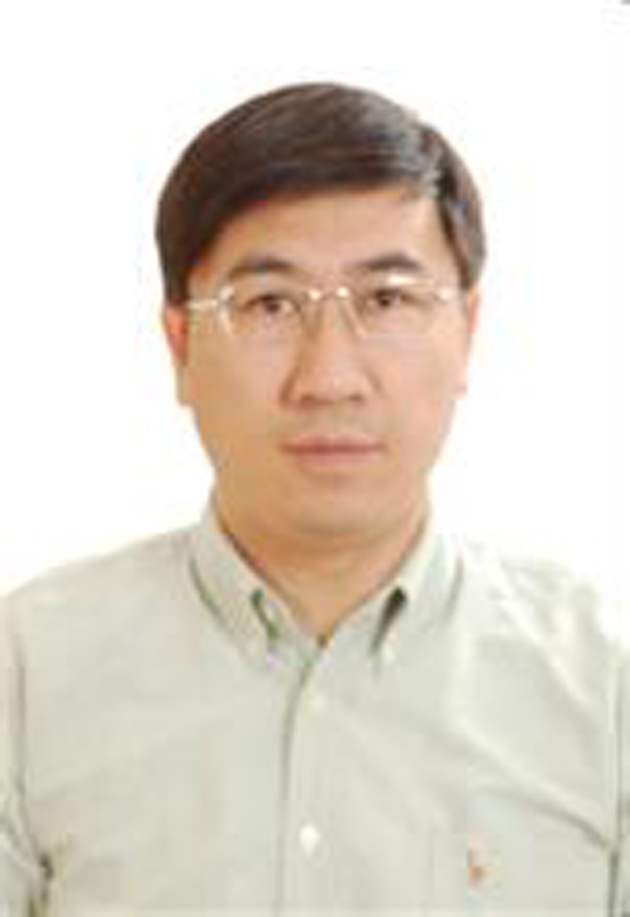}}]{Qingming Huang} received the B.S. degree in computer science and Ph.D. degree in computer engineering from the Harbin Institute of Technology, Harbin, China, in 1988 and 1994, respectively. He is currently a Chair Professor with the School of Computer Science and Technology, University of Chinese Academy of Sciences. He has authored over 400 academic papers in international journals, such as IEEE Transactions on Pattern Analysis and Machine Intelligence, IEEE Transactions on Image Processing, IEEE Transactions on Multimedia, IEEE Transactions on Circuits and Systems for Video Technology, and top level international conferences, including the ACM Multimedia, ICCV, CVPR, ECCV, VLDB, and IJCAI. He is the Associate Editor of IEEE Transactions on Circuits and Systems for Video Technology and the Associate Editor of Acta Automatica Sinica. His research interests include multimedia computing, image/video processing, pattern recognition, and computer vision.
\end{IEEEbiography}



\end{document}